\documentclass[twocolumn,3p,authoryear]{elsarticle}
\usepackage[utf8]{inputenc}
\usepackage[T1]{fontenc}
\usepackage{stix} 
\usepackage{amsmath} 
\usepackage{lineno} 
\usepackage{float} 
\usepackage[breaklinks=true]{hyperref}  
\usepackage[capitalize]{cleveref} 
\usepackage{xurl} 
\usepackage{booktabs}
\usepackage{multirow}
\usepackage{graphicx}
\usepackage{xspace}
\usepackage{dblfloatfix}

\usepackage[table]{xcolor}

\usepackage{caption}
\journal{Remote Sensing of the Environment}
\hypersetup{colorlinks=true} 

\newcommand{\fdelta}{\mbox{FORMSpoT-\texorpdfstring{$\Delta$}{Delta}}\xspace}
\newcommand{\spotsixseven}{SPOT-6/7\xspace}
\newcommand{\formst}{FORMS-T\xspace}
\newcommand{\mtwo}{~m\textsuperscript{2}\xspace}
\begin{document}

\hypersetup{
    citecolor=[rgb]{0.1, 0.3, 0.4},
    linkcolor=[rgb]{0.1, 0.3, 0.4},
    urlcolor=[rgb]{0.1, 0.3, 0.4},
}

\begin{frontmatter}

\title{FORMSpoT: Revealing Fine-Scale Forest Disturbances from Nation-Wide 1.5 m Forest Canopy Height Time Series}

\author[Lsce]{Martin Schwartz} 
\author[ENS]{Fajwel Fogel}
\author[IGNEIF,IGNLOR]{Nikola Besic} 
\author[EcoVision]{Damien Robert}
\author[LIGM]{Louis Geist}
\author[IGNEIF,IGNLOR,ONF]{Jean-Pierre Renaud}
\author[LESSEM]{Jean-Matthieu Monnet}
\author[leipzig]{Clemens Mosig}
\author[IGNEIF,IGNLOR]{Cédric Vega} 
\author[ENS]{Alexandre d'Aspremont}
\author[LIGM]{Loic Landrieu}
\author[Lsce]{Philippe Ciais} 

\affiliation[Lsce]{
  organization={Laboratoire des Sciences du Climat et de l’Environnement, LSCE/IPSL, CEA-CNRS-UVSQ, Université Paris-Saclay},
  city={Gif-sur-Yvette},
  postcode={91191}, 
  country={France}
}
\affiliation[ENS]{
  organization={CNRS \& Département d'Informatique, École Normale Supérieure - PSL},
  addressline={45 Rue d'Ulm},
  postcode={75005}, 
  city={Paris},
  country={France}
}

\affiliation[IGNEIF]{
  organization={Université Gustave Eiffel, Géodata Paris, IGN, LIF},
  city={Nancy},
  postcode={54000}, 
  country={France}
}
\affiliation[IGNLOR]{
  organization={Université de Lorraine, Géodata Paris, IGN, LIF},
  city={Nancy},
  postcode={54000}, 
  country={France}
}
\affiliation[EcoVision]{
  organization={EcoVision Lab, DM3L, University of Zurich},
  country={Switzerland}
}
\affiliation[LIGM]{
  organization={LIGM, École des Ponts, CNRS, IPP, UGE}
}

\affiliation[ONF]{
  organization={Office National des Forêts, Pôle Recherche et Développement Innovation},
  addressline={8 Allée de Longchamp},
  city={Villier-les-Nancy},
  postcode={54600}, 
  country={France}
}
\affiliation[LESSEM]{
  organization={Université Grenoble Alpes, INRAE, LESSEM},
  addressline={2 rue de la Papeterie-BP 76},
  postcode={38402}, 
  city={St-Martin-d'Hères},
  country={France}
}
\affiliation[leipzig]{
  organization={Institute for Earth System Science and Remote Sensing},
  city={Leipzig},
  country={Germany}
}

\begin{abstract}
Current large-scale satellite-based forest disturbance monitoring systems operate at 10-30~m resolution, too coarse to detect changes at the scale of individual trees and resulting in a systematic underestimation of forest disturbances.
Here, we introduce FORMSpoT (Forest Mapping with SPOT Time series), a decade-long (2014-2024), country-scale mapping of forest canopy height at 1.5~m resolution over France, together with \fdelta, annual disturbance polygons derived from height differences in the FORMSpoT time series.

Canopy heights were derived from annual SPOT-6/7 composites using a hierarchical transformer model (PVTv2) trained on high-resolution airborne laser scanning (ALS) data. To enable robust change detection, we developed a post-processing pipeline combining co-registration and spatio-temporal total variation denoising.

We find that (1) the French disturbance regime is dominated by small events. Sub-100\mtwo disturbances alone represent 72\% of all events, and disturbances below 0.1~ha account for 97\% of events and 39\% of the disturbed area. These events are largely missed by Sentinel-1/2 and Landsat-based products. (2) Validated against successive ALS revisits across 19 sites and 5,087 NFI plot revisits, \fdelta provides reliable detection (F1~>~0.8) above 100\mtwo while retaining sensitivity to finer events that coarser products do not capture. (3) At the national scale, \fdelta resolves contrasted disturbance regimes, from clear-cut-dominated dynamics in maritime pine plantations to diffuse, smaller disturbance events in mountain forests, and captures their temporal dynamics, including the salvage-logging signature of the 2017-2022 bark beetle crisis in northeastern France

\end{abstract}

\begin{keyword}
  Deep Learning \sep Forest height \sep Time Series \sep SPOT
  \sep ALS \sep Vision Transformers \sep Forest disturbances, canopy gaps

\end{keyword}

\end{frontmatter}

\section{Introduction}

The European Union (EU) strongly relies on forest carbon sinks to fulfill its international engagement in the Paris Agreement through the European Green Deal, which aims for carbon neutrality by 2050 \citep{migliavacca2025securing, euEuropeanGreenDeal2024}. Yet this commitment is compromised by the recent drop in the EU's forest sink \citep{korosuoRoleForestsEU2023}, partly driven by a decline in forest productivity \citep{tijerin-trivinoForestProductivityDecreases2025, hertzogTurningPointProductivity2025, martinezdelcastilloClimatechangedrivenGrowthDecline2022} and an increase in both anthropogenic and natural disturbances \citep{senfCanopyMortalityHas2018, senfMappingForestDisturbance2020}.

Forest disturbances in Europe span a wide range of spatial scales, from large clearcuts, storms or wildfires covering several hectares down to individual tree mortality or selective logging practices leading to the formation of small canopy gaps. Planned anthropogenic disturbances for wood harvesting account for 82\% of canopy openings across European forests \citep{seidlChangesPlannedUnplanned2024}. Natural disturbances such as wildfires, pest attacks, and windthrows represent the remaining fraction and are projected to intensify under climate change, with associated timber losses expected to nearly double by 2076-2100 \citep{mohrRisingCostDisturbances2025}. Beyond these large-scale events, background mortality \citep{idoate-lacasiaTrendsBackgroundMortality2025}  generates small canopy gaps that, despite occurring at the scale of one to a few individual trees, play a fundamental role in forest ecological processes \citep{muscoloReviewRolesForest2014, krugerGapExpansionDominant2024}. In an unmanaged beech forest in Ukraine, for instance, \citet{hobiGapPatternLargest2015} showed that 98\% of canopy openings were smaller than 200\mtwo, illustrating the importance of fine-scale disturbances in unmanaged forests. Yet, the aggregate contribution of these fine-scale events to the overall area of disturbed forests in Europe remains unquantified due to the current limitations of continental-scale monitoring approaches.

National Forest Inventory (NFI) plots provide accurate, fine-scale observations of forest disturbances, such as individual tree mortality or harvesting, and support national carbon reporting in Europe \citep{blujdeaEUGreenhouseGas2015}. However, as a spatially discontinuous sampling network, they require aggregation over large areas and multiple years to be statistically robust, and do not provide spatially explicit information.
Satellite-based monitoring enables spatial continuity, but existing operational products rely on sensors with a resolution too coarse to detect the smallest disturbances. Landsat-based products such as \citet{hansenHighResolutionGlobalMaps2013}, \citet{turubanovaTreeCanopyExtent2023}, and the European Forest Disturbance Atlas \citep{viana-sotoEuropeanForestDisturbance2025} provide multi-decadal disturbance estimations at 30~m resolution for continental and global scale, enabling a theoretical minimum disturbance detection threshold of 900\mtwo.
Sentinel-1-based systems \citep{vanderwoudeEuropeanForestDisturbance2026a} offer a complementary approach, providing near-real-time disturbance alerts. 
Similarly, Sentinel-2 based products provide disturbance maps at 10~m resolution \citep{mosigSubpixelMappingDisturbance2026, paulsCapturingTemporalDynamics2025}.
However, these products are rarely sensitive to disturbances at exactly 10~m or lower \citep{mosigSubpixelMappingDisturbance2026}. 
With tree crowns rarely exceeding 10~m in diameter \citep{jucker2025global}, these products are therefore limited to disturbances affecting multiple trees.

High resolution (HR) and very high resolution (VHR) optical imagery, with spatial resolutions below 4~m and 1~m respectively \citep{qin3DChangeDetection2016a}, offer the possibility to go beyond this limit. Recent studies have demonstrated the potential of VHR imagery for forest structural mapping at large scales (\citep{tuckerSubcontinentalscaleCarbonStocks2023,brandt2026chmv2,liuOverlookedContributionTrees2023}. However, these studies are based on single-date acquisitions and lack the temporal continuity required for disturbance monitoring. Detecting disturbances from HR and VHR imagery requires not just high spatial resolution at one point in time, but geometrically consistent time series that can reliably assess canopy height changes over time. Such a time series would enable the characterization of the fine-scale disturbance regime of European forests at a spatial detail beyond the reach of current 10-30~m monitoring systems.

France offers a well suited to address this gap thanks to two complementary open datasets. First, the DINAMIS initiative \citep{AccesLOpenData} has provided, since 2014, annual country-wide \spotsixseven mosaics at 1.5~m panchromatic resolution, forming an eleven-year HR archive over the largest country in the European Union (551,695~km\textsuperscript{2}). Second, the French LiDAR HD program \citep{ignLiDARHDGeoservices2025} has acquired laser scanning data across the entire territory giving a precise view of forest 3D structure. This data combination is valuable, and French forests span a wide range of species compositions, management regimes, and terrain conditions, making the country a demanding and representative test case for HR-based disturbance monitoring in European temperate forests.

Here, we present FORMSpoT (FORest Mapping with SPOT Time-series), a dataset of annual forest canopy height maps at 1.5~m resolution covering metropolitan France from 2014 to 2024, derived from \spotsixseven imagery using a hierarchical transformer model trained on LiDAR HD data. To ensure temporal consistency across images acquired under varying viewing angles, dates, and atmospheric conditions, we introduce a post-processing pipeline combining spatial co-registration and spatio-temporal total variation denoising. From the resulting height time series, we extract \fdelta, a database of annual fine-scale forest disturbance polygons, validated against successive airborne lidar acquisitions across 19 sites and 5,087 French National Forest Inventory plot revisits, and compared with seven existing coarser resolution disturbance products. We address three research questions: \textbf{(1) What is the effective spatial resolution required to capture the full spectrum of forest disturbances in European temperate forests, and what fraction of disturbances remains invisible to current 10-30~m products?} Using \fdelta, we quantify the size distribution of disturbances across France and, using repeated ALS acquisitions, we evaluate the fraction that falls below the detection limit of Landsat- and Sentinel-based products. \textbf{(2) Can a national-scale HR satellite time series provide reliable annual disturbance monitoring down to the scale of small canopy gaps?} We answer this through a validation of FORMSpoT and \fdelta against successive ALS acquisitions and NFI plot revisits, establishing the size threshold above which detection is reliable. \textbf{(3) How do fine-scale disturbance patterns vary across forest types and regions in France?} We map national disturbance regimes from clear-cut-dominated plantations to diffuse mountain events, and track their temporal dynamics, including the 2017-2022 bark beetle crisis in northeastern France. 
Both FORMSpoT and \fdelta are openly available at \url{https://doi.theia.data-terra.org/FORMSpoT/}.

\section{Study area}
\subsection{French forests}
Metropolitan France (551,695~km\textsuperscript{2}) hosts approximately 17 million hectares of forest (31\% of national territory), three-quarters of which are privately owned. Broadleaf species dominate two-thirds of this area, with oaks and beech prevailing across northern and western plains, while evergreen oaks and pines characterize the Mediterranean basin. Conifer forests of fir, spruce and larch occupy the mountainous region of the Alps, Vosges, Jura and Pyrenees, and the vast Landes maritime pine plantation in the southwest forms the largest planted forest in Europe. Disturbances can vary greatly in size and shape, ranging from single-tree mortality to multi-hectare events resulting from harvesting or natural hazards such as large windthrows or wildfires.

\subsection{Test sites}
\label{sec:successive_als_sites}

\begin{figure}[t]
  \centering
  \includegraphics[width=0.4\textwidth]{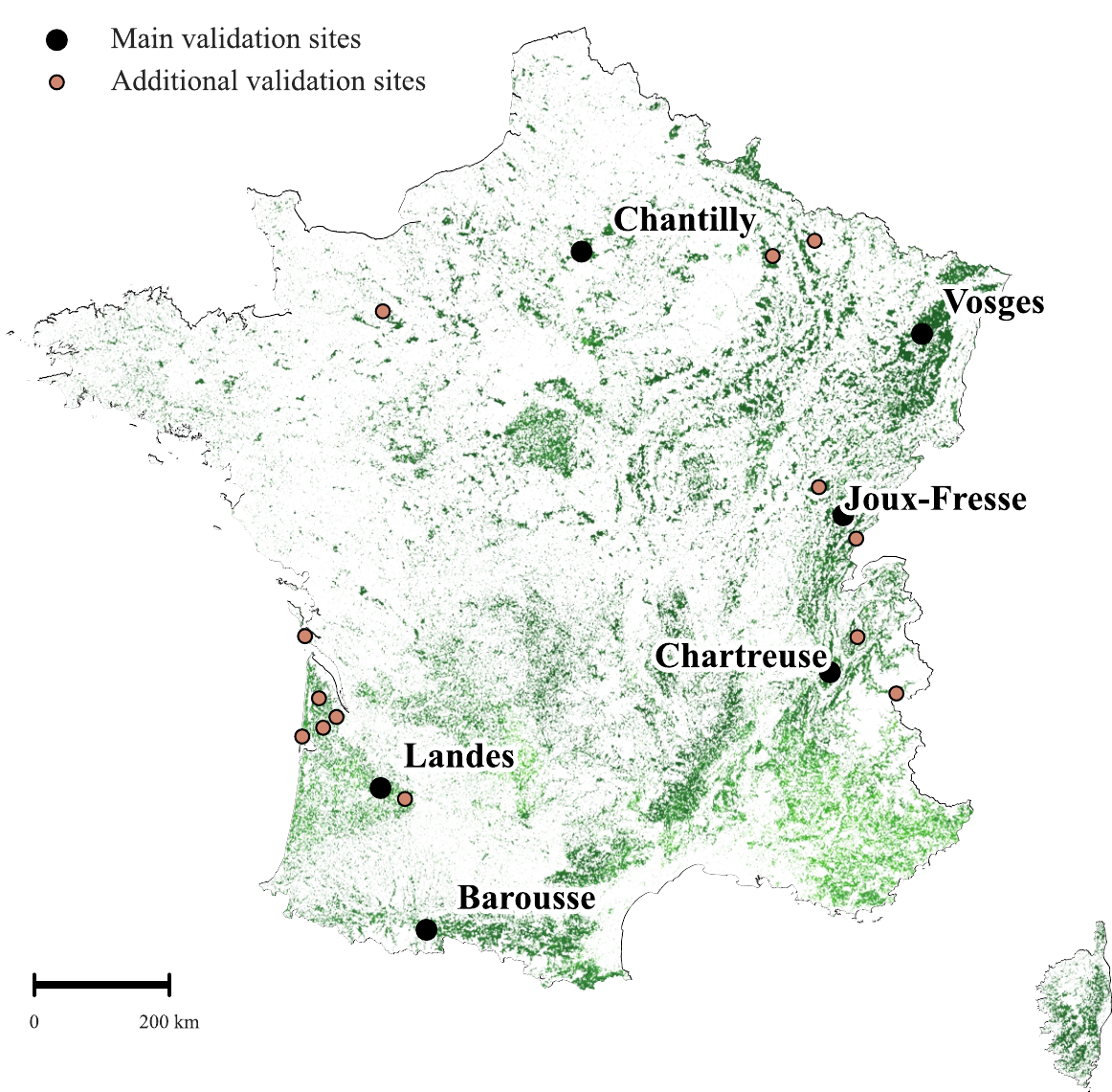}
  \caption{\textbf{Locations of the 19 ALS test sites.} The six representative sites used for the site-level analyses are highlighted in bold black font. Additional test sites (presented in the supplementary Section~1), are indicated with smaller orange dots. Greenness represents tree cover density \citep{clmsTreeCoverDensity2023}.}
  \label{fig:2-ALS_sites}
\end{figure}

We selected 19 sites distributed across France (\cref{fig:2-ALS_sites}) based on their availability of two ALS acquisitions at least one year apart, that we used to validate FORMSpoT height, tree cover (\cref{sec:height_tc_validation}) and disturbance detection (\cref{sec:methods:grid_based_tc_change}). All 19 sites contribute to the validation of FORMSpoT height, tree cover, and disturbance detection. For the site-level analyses and detailed comparisons (\cref{fig:results:viz_fdelta-ljlf_barousse,fig:6-visual_comparison_sota,fig:7-recall_precision_joux_fresse,fig:8-F1_scores}), we present six representative sites spanning a broad range of forest and terrain configurations (\cref{tab:1_ALS-sites}). These include high mountainous regions, where detecting disturbances is particularly challenging (e.g., Chartreuse, Barousse); lower mountains with mixed forests (e.g., Vosges, Joux-Fresse); and low-relief areas such as an intensively managed maritime pine plantation (Landes) and a declining oak forest (Chantilly). In Chantilly, salvage logging of declining oaks creates sparse canopy gaps. The 13 additional sites are presented in the Supplementary Materials of this study.

\begin{table*}[t]
  \centering
  \caption{\textbf{Description of the 6 representative sites.} Site locations are presented in \cref{fig:2-ALS_sites}. The description and results for additional test sites can be found in the supplementary materials.}
  \renewcommand{\arraystretch}{1}
\scriptsize
\begin{tabular}{p{1.5cm} p{3.3cm} p{3.8cm} p{4.4cm} p{1.3cm}}
    \multicolumn{1}{c}{\textbf{Site name}} & \multicolumn{1}{c}{\textbf{Forest type}} & \multicolumn{1}{c}{\textbf{ALS year 1}} & \multicolumn{1}{c}{\textbf{ALS year 2}} & \multicolumn{1}{c}{\textbf{Area}}
    \\
    \toprule
    \centering Landes \vspace{3pt} \newline  \includegraphics[width=0.7cm]{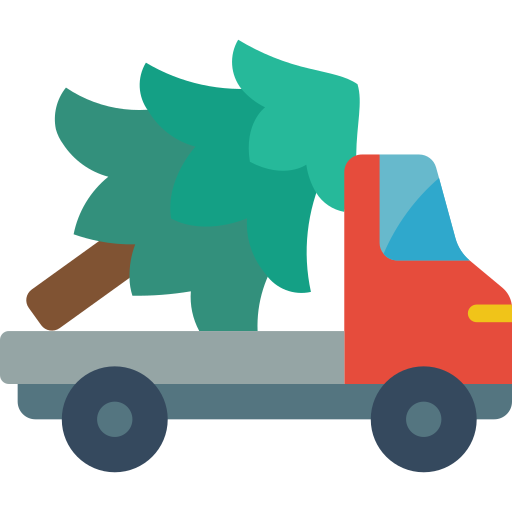} &
    Maritime pine plantations around the Ciron river &
    Oct. 2019 - FRISBEE research project \citep{durrieuDonneesLIDARULM2025a} &
    March 2021 - FRISBEE research project \citep{durrieuDonneesLIDARULM2025} &
    $\sim$1,500~ha \\
    \midrule
    
    \centering Chantilly \vspace{3pt} \newline  \includegraphics[width=0.7cm]{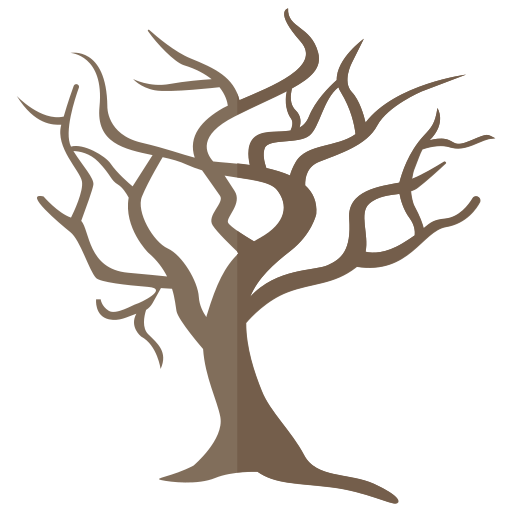} &
    Declining deciduous oak forest with salvage logging over a flat area \citep{leboulerEnsembleSauvonsForet2023} &
    Feb. 2022 - Lidar HD, IGN &
    Sep. 2023 - Requested to Laurent Saint-André and the collective ``Ensemble Sauvons la forêt de Chantilly.'' &
    $\sim$7,000~ha \\
    \midrule
    
    \centering Joux-Fresse \vspace{3pt} \newline  \includegraphics[width=0.7cm]{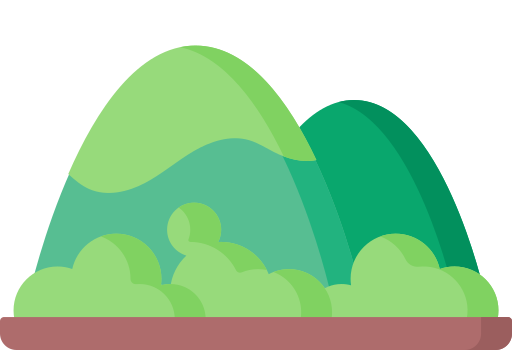} &
    Coniferous in low mountains (Jura) &
    June 2019 - Office National des Forêts (ONF) &
    Aug. 2022 - Lidar HD, IGN &
    $\sim$6,000~ha \\
    \midrule
    
    \centering Vosges \vspace{3pt} \newline  \includegraphics[width=0.7cm]{Images/mountain_small.png} &
    Mixed forests (beech, fir, spruce) in low mountainous areas (Vosges) &
    Feb. 2020 - IGN, see \citep{ramirezparraHowReliableAre2024} for description &
    Apr. 2022 - Lidar HD, IGN &
    $\sim$225,000~ha \\
    \midrule
    
    \centering Chartreuse \vspace{3pt} \newline  \includegraphics[width=0.7cm]{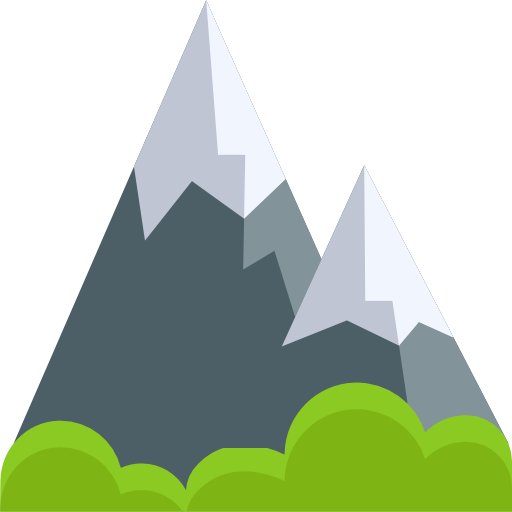} &
    Mixed forest, in high mountainous areas (Alps) &
    Sep. 2016 - Office National des Forêts (ONF) &
    Sep. 2021 - Lidar HD, IGN &
    $\sim$9,000~ha \\
    \midrule
    
    \centering Barousse \vspace{3pt} \newline  \includegraphics[width=0.7cm]{Images/mountain.png} &
    Mixed forest, in high mountainous areas (Pyrenees) &
    Sep. 2019 - Office National des Forêts (ONF) &
    Sep. 2021 - Lidar HD, IGN &
    $\sim$5,200~ha \\
    \bottomrule

\end{tabular}
  \label{tab:1_ALS-sites}
\end{table*}

\section{Data}

\subsection{Training data: The Open-Canopy benchmark dataset}
\label{sec:methods:opencanopy}
Open-Canopy \citep{Fogel_2025_CVPR} is an open-access benchmark dataset for canopy height estimation, covering 87,383~km\textsuperscript{2} of mainland France (2021-2023). It pairs \spotsixseven pansharpened images at 1.5~m resolution with canopy height models derived from the French LiDAR HD program \citep{ignLiDARHDGeoservices2025}. The dataset spans diverse acquisition conditions and is split into training (66,339 km\textsuperscript{2}), validation (7,369 km\textsuperscript{2}), and test (13,675 km\textsuperscript{2}) sets stratified across bioclimatic regions, with spatial buffers to prevent leakage. Full details are provided in \citet{Fogel_2025_CVPR}. In this study, we used the pretrained model weights released with the dataset (see \cref{sec:methods:model_training_inference}). Test sites overlapping with Open-Canopy training or validation splits were excluded from all analyses.

\subsection{Annual \spotsixseven mosaics for multi-year national mapping (2014-2024)}
\label{sec:spot_data}
The \spotsixseven constellation comprises two CNES satellites (launched in 2012 and 2014) acquiring optical imagery in four spectral bands at 6~m and one panchromatic band at 1.5~m. Country-wide orthorectified mosaics over France (2014-2024) are freely distributed for international users annually through the DINAMIS platform \citep{AccesLOpenData}. Each annual mosaic is assembled from 240 images covering 60~km $\times$ 60~km tiles over metropolitan France and Corsica, selected by DINAMIS from 400-600 \spotsixseven acquisitions made between April and September of each year according to specifications on viewing angles, cloud cover and snow cover defined by IGN \citep{dinamisCouverturesNationalesSPOT2020, applisatCouvertureNationaleSpot}. We retrieved the multispectral and panchromatic bands and applied Brovey pansharpening \citep{gdal/ogrcontributorsGDALOGRGeospatial2025} to produce multispectral data at 1.5~m resolution, hereafter referred to as "\spotsixseven images".

\subsection{Validation}
\subsubsection{ALS validation data}
\label{sec:methods:als_data}

We gathered 38 high-resolution (\textgreater 5 pts/m\textsuperscript{2}) ALS acquisitions from various sources detailed in Table~S1, with two acquisitions per site to enable change detection between revisits.
Half of the validation sites originate from the French LiDAR HD program, a nationwide ALS survey launched in 2021. Because LiDAR HD is also the source of the Open-Canopy dataset used to train our model (\cref{sec:methods:opencanopy}), particular care was taken to avoid any spatial overlap between training and validation. Validation sites are either entirely new locations not present in the training set, or, where a geographic intersection existed, the overlapping area was removed from the validation data so that no validation pixel contributed to model training.
We used published canopy height models (CHMs) at 0.5~m resolution \citep{ignLiDARHDGeoservices2025} for LiDAR HD data, and 1~m resolution CHMs provided by other data providers. All CHMs were reprojected to 1.5~m resolution aligned with the \spotsixseven grid, taking maximum values in each pixel. These datasets are hereafter referred to as "ALS CHMs".

\subsubsection{French NFI plots}

\begin{figure*}[b]
  \centering
  \includegraphics[width=\textwidth]{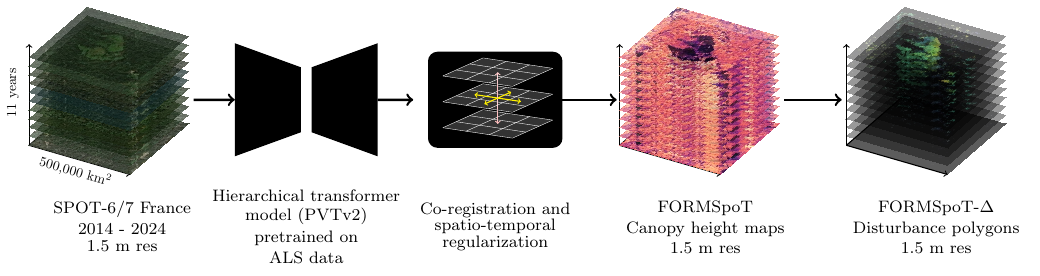}
  \caption{\textbf{Overview of the FORMSpoT and \fdelta generation workflow.} SPOT-6/7 images over mainland France (1.5~m resolution, 2014-2024) were processed with a pretrained PVTv2 model \citep{Fogel_2025_CVPR} to produce annual forest canopy height maps, which were then co-registered and spatio-temporally regularized. From the resulting height time series (FORMSpoT), disturbance polygons (\fdelta) were extracted using a 5~m height-loss threshold and a morphological filter.}
  \label{fig:1-Workflow}
\end{figure*}

\label{sec:NFI_data}
The French National Forest Inventory (NFI) collects measurements from over 10,000 field plots annually across the country, recording dendrometric attributes (e.g., tree height, diameter) and environmental indicators within 30~m diameter areas. To validate our FORMSpoT height estimates, we compared the tallest tree height in each NFI plot with the maximum FORMSpoT pixel value within an 18~m radius around the plot center, accounting for plot extent and a 3~m geolocation error buffer. Over 2014-2024, this yielded 59,371 plots, of which 8,337 were excluded as they fell within model training areas. Acquisition times were matched by comparing each NFI plot measurement with the FORMSpoT map of the corresponding year. 
Since 2010, all plots have been re-measured (diameters only) on a five-year cycle \citep{bontempsTakeFiveBeat2024}.
We used these revisits to compute the percentage of trees that disappeared within the 2018-2023 interval and compared these values with our predicted disturbance polygons (\fdelta, \cref{sec:methods:fdelta_polygons}), yielding 5,087 plots, of which 679 from training areas were excluded. 

\subsection{Coarser resolution disturbance datasets used for comparison}
\label{sec:sota_data}
To evaluate the benefit of HR imagery, we compared our disturbance detection results with previously published datasets. Some rely on Landsat imagery at 30~m resolution: (1) \citet{hansenHighResolutionGlobalMaps2013}, (2) \citet{turubanovaTreeCanopyExtent2023}, and (3) the European Forest Disturbance Atlas (EFDA; \citet{viana-sotoEuropeanForestDisturbance2025}). Others are derived from Sentinel-1 and/or Sentinel-2: (4) SUFOSAT \citep{mermozSubmonthlyAssessmentTemperate2024}, (5) \formst \citep{schwartzRetrievingYearlyForest2025}, (6) \citet{paulsCapturingTemporalDynamics2025}, and (7) the yearly deadtrees.earth tree cover map excluding its standing deadwood \citep{mosigSubpixelMappingDisturbance2026,mosigDeadtreesearthOpenaccessInteractive2026}. The latter three do not explicitly provide disturbance estimates; we derived them using a 5~m height-loss threshold (\formst and Pauls) consistent with \fdelta (\cref{sec:methods:fdelta_polygons}), and a 25\% tree cover loss threshold for deadtrees.earth, selected as the value maximizing F1 score across all sites (\cref{sec:methods:grid_based_tc_change}). A detailed description of these products is provided in Table~S2. Additionally, we used a forest mask \citep{ignMasqueForetV12025} to compute the forest fraction in \cref{sec:results:spatio_temporal_patterns}.

\section{Methods}
\label{sec:methods}

\subsection{Height model and inference}
\label{sec:methods:model_training_inference}
The architectural choice and its benchmarking against alternative models for canopy height estimation from \spotsixseven imagery and ALS data are discussed in detail in \citet{Fogel_2025_CVPR}. Building on that work, we focus here on the nationwide, decade-long application and on the post-processing required to obtain a consistent multi-year time series. We used the PVTv2 pyramid transformer model \citep{wangPVTV2Improved2022}, pretrained on the Open-Canopy dataset (\cref{sec:methods:opencanopy}). The PVTv2 architecture processes the image through successive stages of decreasing spatial resolution, so that features are extracted at several scales relevant for canopy height estimation: fine enough to represent local crown texture, and coarse enough to encode the surrounding context, such as whether a pixel lies within a dense stand, near a forest edge, or in an open area. Architecture details and model weights are available in \citet{wangPVTV2Improved2022} and \citet{Fogel_2025_CVPR}, respectively. As described in \citet{Fogel_2025_CVPR}, the model was trained on \spotsixseven mosaics from 2021-2023, spanning a wide range of acquisition conditions (seasonality, atmospheric conditions, sun angle), making it suitable for generalization to other years. We applied it to all overlapping \spotsixseven images covering France for 2014-2024 (\cref{sec:spot_data}), averaging overlapping regions to obtain a single country-wide prediction per year. The mathematical formalism is provided in supplementary Section~3.1.

\subsection{Post-processing of the height time series}

The high variability in \spotsixseven images due to different acquisition conditions (angle, time of the year, aerosols) and registration errors introduces temporal and spatial inconsistencies in the predicted dataset that we addressed by applying the post-processing steps described below. We refer to the resulting height time series as FORMSpoT (FORest Mapping with SPOT Time series).

\subsubsection{Co-registration}
\label{sec:coregistration}
To correct for small spatial misalignments between annual predictions, we aligned each predicted height map to the predicted 2019 height map, as it represented the middle of the time series. 
To restrict the magnitude of the registration, we only allowed a shift of two pixels in each direction. Then we chose the shift that minimized the squared difference between the reference (year 2019) and the shifted height map (other years). We applied this registration procedure by patches of 2000~x~2000 pixels. The patch size is large enough that even in the presence of disturbances, the registration remains driven by the dominant stable canopy structure, since localized changes affect only a small fraction of each patch. These patches were taken with an overlap ensuring that, after registration, each input pixel remained covered by at least one shifted patch. Registered predictions in the overlapping regions were then averaged to produce a seamless, registered time series of canopy height maps.  
Performing this operation after the model inference is easier due to the highly heterogeneous reflectance ranges in \spotsixseven images, which would make the co-registration algorithm more complicated to apply.
The mathematical formalism for this operation can be found in the supplementary Section 3.2.

\subsubsection{Total variation (TV) denoising}
\label{sec:TV_denoising}
The various conditions of acquisition in \spotsixseven images introduce a temporal noise in our height predictions. However, forest canopy height naturally evolves in a structured way with slow increase over time due to tree growth, whereas decreases are abrupt and associated with discrete events such as cuts, storms or wildfires. Based on this prior, we applied a spatio-temporal total variation (TV) denoising scheme to our time-series using the Chambolle-Pock primal-dual algorithm \citep{chambolleFirstOrderPrimalDualAlgorithm2011} that we re-implemented within a GPU-compatible framework to enable efficient large-scale processing. The TV regularization increases spatial coherence and forms temporally smooth, piecewise-constant trajectories while preserving the sharp height drops that correspond to real disturbances. This regularization is parameterized by spatial $\lambda_{\text{spat}}$ and temporal $\lambda_{\text{temp}}$  weights that affect the strength of the denoising in each dimension.  We empirically set $\lambda_{\text{temp}} = 5 $ and $\lambda_{\text{spat}} = 0.5$ in this study. The temporal regularization is higher because stable height trajectories are essential to detect disturbances, while spatial regularization had only a minor effect and was used mainly to mitigate tiling artifacts. The mathematical formalism for this section can be found in the supplementary Section 3.3, and the effect of TV denoising is illustrated in Appendix \cref{fig:11-TV_effect}. 

\subsubsection{\fdelta polygons}
\label{sec:methods:fdelta_polygons}
From the FORMSpoT height maps, we derived \fdelta, a dataset providing all detected disturbance polygons covering mainland France for the 2014-2024 period. 
For this, we defined a disturbance as a negative change of more than 5~m in height at the pixel level. In the result part of this study (\cref{sec:results:fdelta}), we carried out a sensitivity analysis to assess the effects of this threshold on the disturbance detection capacity and showed that 5~m maintains the best balance between precision and recall across all disturbance sizes.  
We removed small artifacts and regularized mask contours using an opening morphological filter \citep{serraImageAnalysisMathematical1982} with a kernel size of 3 pixels, which set the minimum disturbance detection threshold to 20\mtwo (9 pixels). A \fdelta polygon constitutes here a proxy for an ecologically meaningful disturbance event. However a single clearing may occasionally be split into several polygons or adjacent fellings occurring in the same year may merge into one.  

\subsection{Height and tree cover validation}
\label{sec:height_tc_validation}
\paragraph{Height validation} We compared our estimations to the 38 ALS CHMs (See \cref{sec:methods:als_data}) over all sites by comparing each CHM with the FORMSpoT map of the corresponding year. We also compared FORMSpoT heights to the French NFI plots (See \cref{sec:NFI_data}) by comparing the maximum FORMSpoT height over the plot to the highest measured tree in the plot. 
\paragraph{Tree cover validation} We applied a forest/non-forest threshold of 5 m to simplify the definition of a forest \citep{faoNationalForestInventory} and create a binary mask for both FORMSpoT maps and the ALS CHMs over all sites that we then aggregated at 30 m to obtain a tree cover map. 

\subsection{Disturbances validation}
\subsubsection{Gridded disturbances with ALS CHMs}
\label{sec:methods:grid_based_tc_change}
\begin{figure*}[t]
    \centering
    \includegraphics[width=\linewidth]{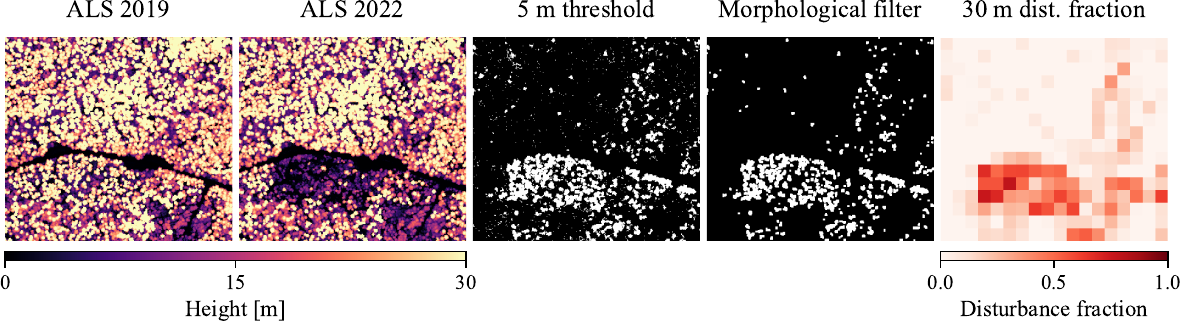}
    \caption{\textbf{Steps to compute the grid-based disturbance fraction datasets.} For each site, a 5~m height-loss threshold is applied to the difference of ALS CHMs to generate a binary disturbance mask. This mask undergoes a cleaning operation using an opening morphological filter \citep{serraImageAnalysisMathematical1982} with a kernel size of 3. It is then downsampled on a 30~m grid to obtain a reference disturbance fraction dataset (here for the Joux-Fresse site, at 46.83\textdegree{}N, 6.02\textdegree{}E).}
    \label{fig:methods:lidar_morpho_grid_based}
\end{figure*}

We performed a 30~m grid-based evaluation of disturbances to enable comparison with coarser-resolution products. For each site, a reference binary disturbance mask was computed by applying a 5~m height-loss threshold to the difference of the two ALS CHMs, followed by the same opening morphological filter used for \fdelta (\cref{sec:methods:fdelta_polygons}). This mask was then aggregated to a 30~m grid to obtain a reference disturbance fraction dataset (\cref{fig:methods:lidar_morpho_grid_based}).

We then aggregated \fdelta and the other disturbance datasets (\cref{sec:sota_data}) to the same 30~m grid, applying the same 5~m threshold and opening filter for \fdelta. For other datasets, disturbance estimates were reprojected to the 30~m grid, yielding binary values for 30~m resolution products and fractional values ranging from $\frac{1}{9}$ to 1 for 10~m resolution products. A 30~m cell was considered disturbed whenever its disturbed fraction was strictly positive, for both the reference and the evaluated products. We computed precision, recall and F1-score for all disturbances, and per disturbance size class. For recall, the size class was computed using the ALS disturbed fraction. For precision, the size class was computed using the product's own disturbed fraction. Consequently, 30~m products are evaluated for precision only at the whole-pixel scale, since their disturbed fraction can only be 0 or 1. The details for precision, recall, and F1-score calculation can be found in  \ref{sec:appendix:metrics_f1_precision_recall_gridbased}.

\subsubsection{Disturbances from NFI plots}
In addition to the ALS-based evaluation, we compared \fdelta with disturbance information from the 2018-2023 French NFI plot revisits. Trees present during the first visit are re-measured during the second visit (five years later), and their status is recorded. A plot was defined as disturbed if at least one tree was cut or fallen during the 2018-2023 period. Standing dead trees were counted as undisturbed, as they do not represent actual canopy height loss, though their presence may influence height predictions since the model was trained on ALS data sensitive to standing dead trees \citep{heinaroComparingFieldMeasurements2023}. New trees passing the measurement threshold (ingrowth) are recorded separately and do not affect our disturbance metric which is based on the disappearance of previously measured trees. The results of the comparison with each NFI plot (5,087) were assigned to one of four categories: true positive (TP): disturbed plot intersecting a \fdelta polygon with disturbance arising during 2018-2023; true negative (TN): undisturbed plot with no intersection; false positive (FP): undisturbed plot intersecting a polygon; false negative (FN): disturbed plot with no intersection. Precision ($P$), recall ($R$), and F1-score ($F1$) were computed as:
\begin{equation}
    P = \frac{TP}{TP + FP}, \quad R = \frac{TP}{TP + FN}, \quad F1 = \frac{2 \times P \times R}{P + R}
\end{equation}
We also computed recall as a function of disturbance severity, defined as the percentage of trees lost between 2018 and 2023 (100\% corresponding to a clear-cut). Alternative formulations based on basal area loss and on larger trees only (diameter at breast height greater than the plot-level mean) yielded similar patterns, so the simpler tree-count metric was retained. The same analysis was applied to other disturbance products covering the 2018-2023 period (Hansen, EFDA, \formst, SUFOSAT).
NFI plot revisits provide a complementary perspective to the ALS-based validation by recording individual tree removals regardless of their spatial configuration. It includes scattered mortality and low-intensity harvesting events like thinning that may leave no canopy gap detectable even by airborne lidar. This makes the NFI comparison a demanding test of \fdelta sensitivity to the smallest disturbance events.

\section{Results}

\subsection{FORMSpoT: National-scale canopy height time series}
\subsubsection{Overview of the dataset}

\begin{figure*}[t]
  \centering
  \includegraphics[width=\textwidth]{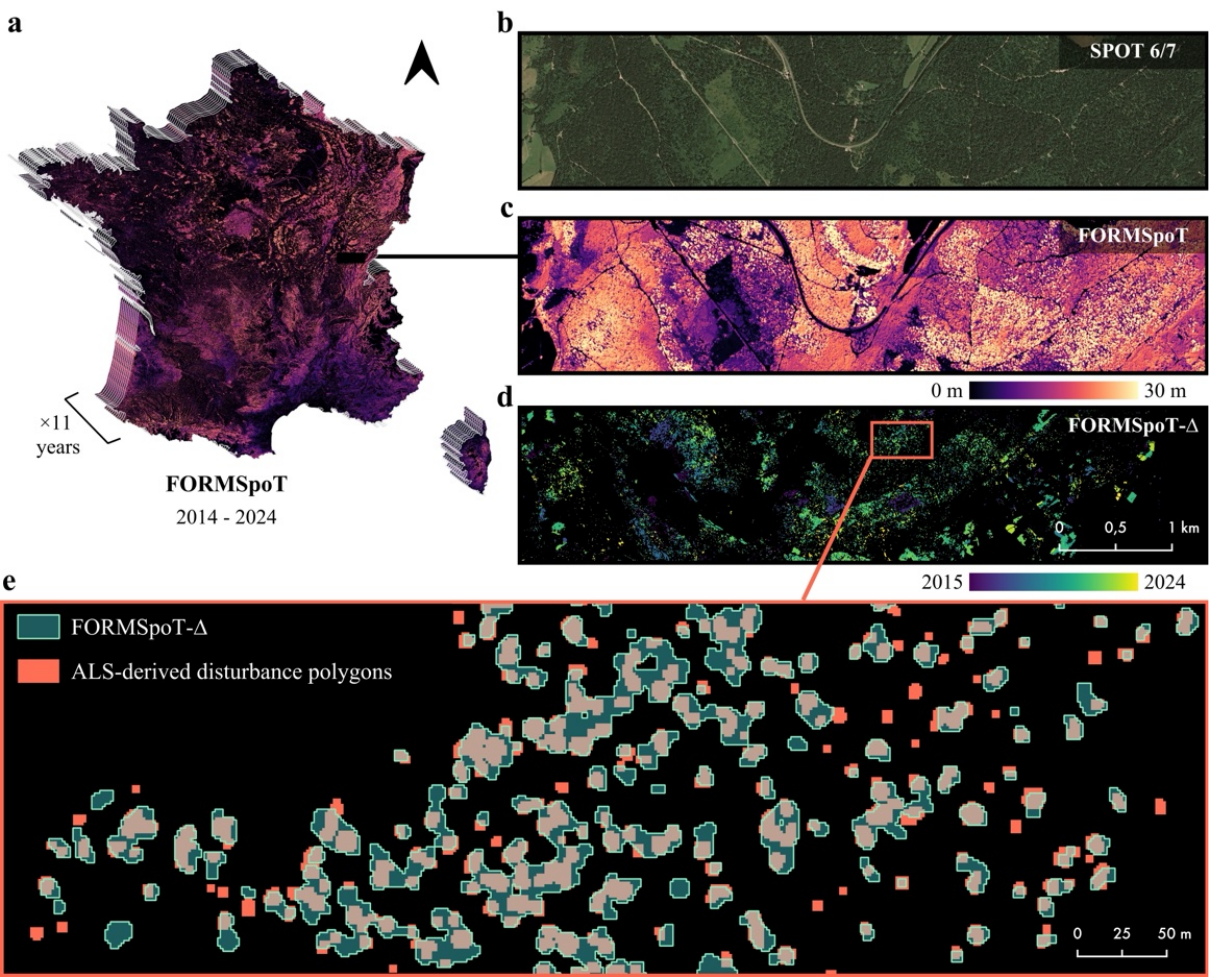}
  \caption{\textbf{FORMSpoT height maps and \fdelta polygons.} a) FORMSpoT maps at 1.5~m resolution. b) Closer view of a \spotsixseven image (2022) over the Jura mountains. c) FORMSpoT predicted height (2022) at 1.5~m resolution over this area. Brighter colors indicate higher heights. d) \fdelta disturbance polygons derived from the FORMSpoT height time series over this area. Brighter colors indicate more recent forest disturbances. e) Comparison of disturbance polygons derived from two ALS flights (2019-2022) with \fdelta polygons spanning these years.}
  \label{fig:4-Qualitative_results}
\end{figure*}

\cref{fig:4-Qualitative_results} illustrates the dataset at multiple spatial scales: a national overview of the 2022 height map, a site-level zoom over the Jura mountains, and a comparison between \fdelta polygons and ALS-derived reference disturbances at the same location. Both FORMSpoT and \fdelta times series can be accessed at \url{https://doi.theia.data-terra.org/FORMSpoT/}.

\subsubsection{Height and tree cover validation}
\begin{figure*}[t]
  \centering
  \includegraphics[width=0.9\textwidth]{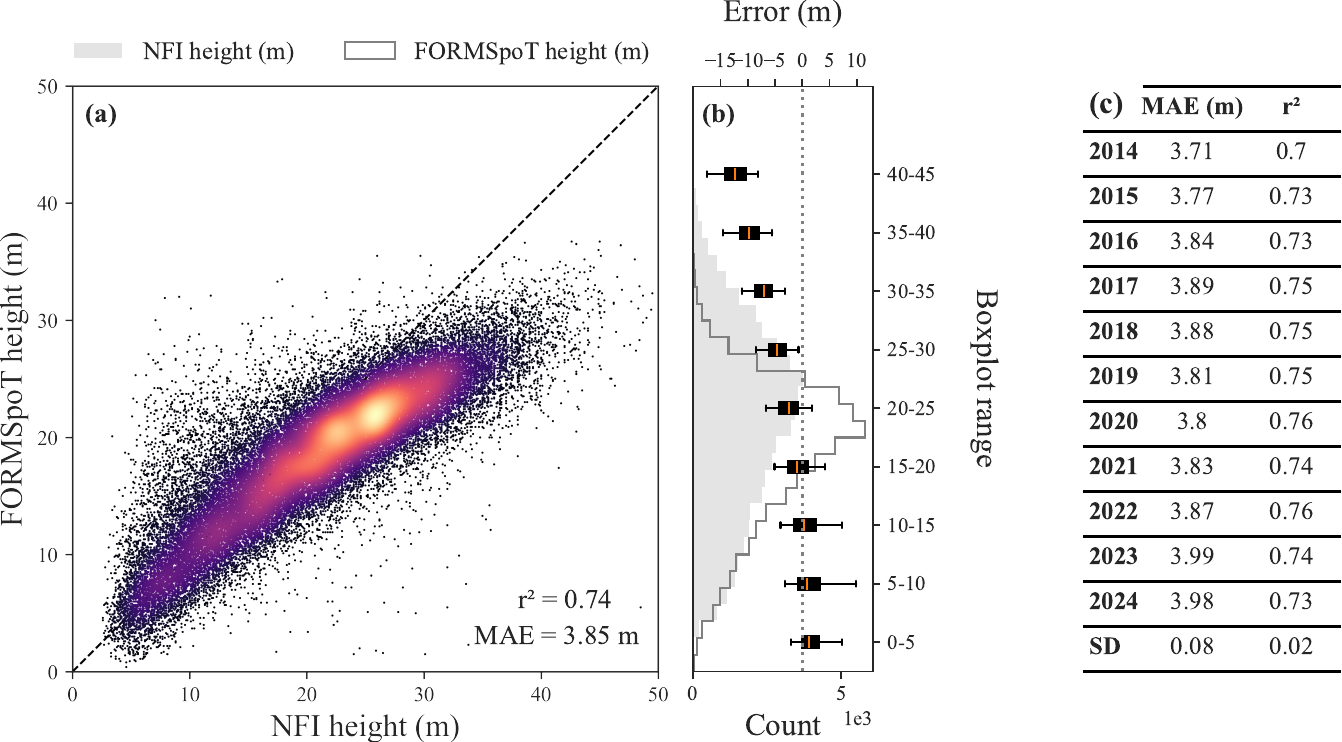}
  \caption{\textbf{Validation of FORMSpoT canopy height estimates using NFI plots across 11 years.} (a) Comparison of FORMSpoT and NFI heights (\cref{sec:NFI_data}), with brighter colors indicating higher point density. The dashed line represents the 1:1 relationship. (b) Distribution of height differences within 5~m reference height bins; the red line denotes the median, box edges indicate quartiles, and whiskers the 5th-95th percentiles. (c) MAE and $r^{2}$ per year (2014-2024), with standard deviation across years in the last row.}
  \label{fig:5-NFI_static_validation}
\end{figure*}

\begin{figure}[H]
    \centering
    \includegraphics[width=\linewidth]{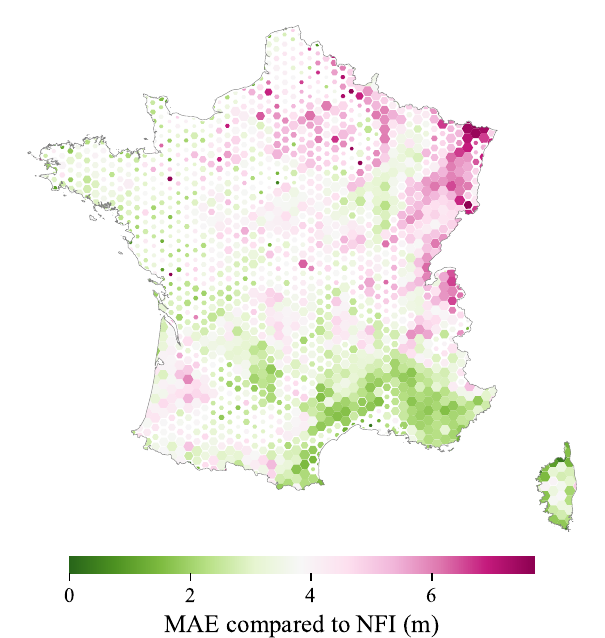}
    \caption{\textbf{Spatial distribution of FORMSpoT height errors across France.} MAE per hexagon between FORMSpoT and NFI height measurements, aggregated over 2014-2024. Green hexagons indicate areas below the national average MAE of 3.85~m. Hexagon size scales with forest fraction and is shown at full size when forest fraction exceeds 40\%.}
    \label{fig:validation_per_hexagon_NFI}
\end{figure}

Against NFI reference heights, FORMSpoT achieves an $r^{2}$ of 0.74 and a MAE of 3.85~m (\cref{fig:5-NFI_static_validation}a); against ALS CHMs, a MAE of 3.54~m and $r^{2}$ of 0.68 (per-site scatter plots in Supplementary Figure~S2).
Errors are not uniform across height classes (\cref{fig:5-NFI_static_validation}b). Low-to-medium canopies are well reproduced, with a small bias below 20~m (e.g., 0.7~m in the 10-15~m range). Taller forests, above 25 m, show a systematic negative bias increasing with height (e.g., $-$9.8~m in the 35-40~m range). 
Interannual stability is critical for disturbance detection, as false height variations could generate wrong disturbance signals. Despite variable acquisition conditions, FORMSpoT remains consistent over time: yearly MAE shows limited variability (SD~=~0.08~m, range: 3.71-3.99~m) and $r^{2}$ remains stable across years (SD~=~0.02, range: 0.70-0.76).
Errors follow a coherent geographic pattern (\cref{fig:validation_per_hexagon_NFI}): largest in mountainous regions with tall canopies (Vosges, Alps), and lowest in the Mediterranean basin and Corsica, where vegetation is lower and more open, consistent with the ALS-based site-level validation (Supplementary Figure~S2).

FORMSpoT 30~m aggregated tree cover estimates agree well with ALS-derived tree cover across all sites (MAE~=~0.09, $r^{2}$~=~0.83), with MAE ranging from 0.03 (0.9-1 class) to 0.245 (0.5-0.6 class), though tree cover is systematically overestimated at high values. Full tree cover validation details are presented in \ref{sec:appendix:tc_validation}.

\subsection{\fdelta: Disturbance detection across spatial scales}
\label{sec:results:fdelta}
\begin{figure*}[t]
    \centering
    \includegraphics[width=\linewidth]{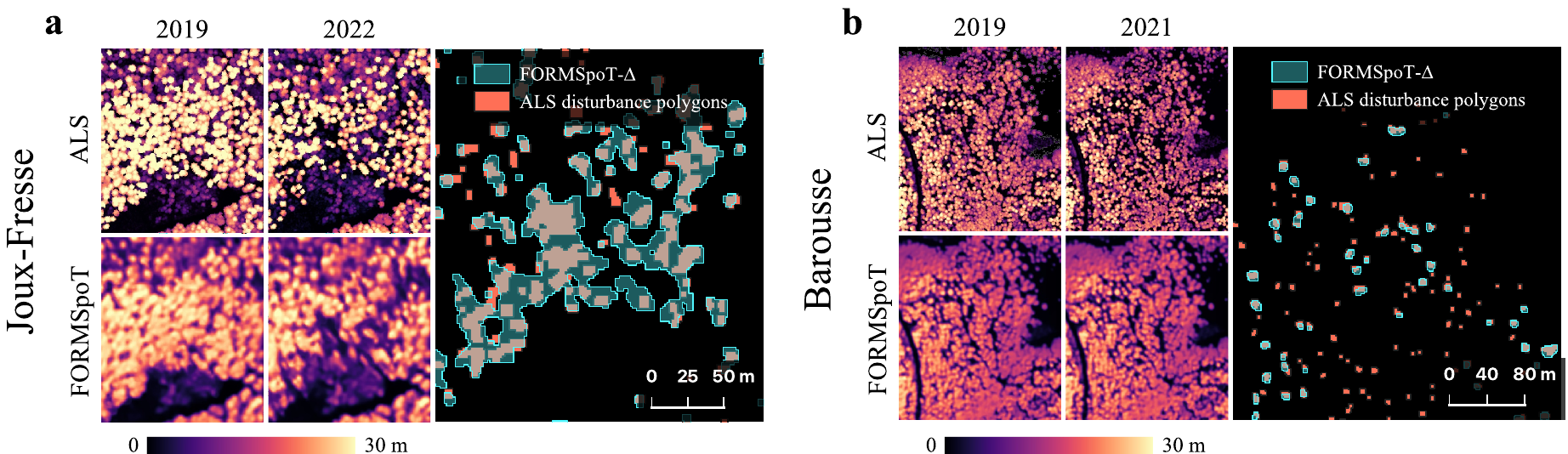}
    \caption{\textbf{Visual comparison of \fdelta and ALS-derived disturbance polygons at two contrasting sites.} (a) Joux-Fresse (46.9\textdegree{}N, 6.0\textdegree{}E), a low-mountain mixed forest with moderate terrain complexity. (b) Barousse (42.9\textdegree{}N, 0.43\textdegree{}E), a high-mountain site dominated by small, spatially fragmented disturbances. ALS-derived polygons serve as reference in both panels.}
    \label{fig:results:viz_fdelta-ljlf_barousse}
\end{figure*}

\cref{fig:results:viz_fdelta-ljlf_barousse} illustrates \fdelta polygons alongside ALS-derived reference disturbances at two contrasting sites. At Joux-Fresse, a low-mountain mixed forest, \fdelta polygons closely match the ALS reference. At Barousse, a high-mountain site with predominantly small and spatially fragmented disturbances, a substantial fraction of small disturbances visible in the ALS CHM are missed by \fdelta.

\begin{figure*}[t]
    \centering
    \includegraphics[width=\linewidth]{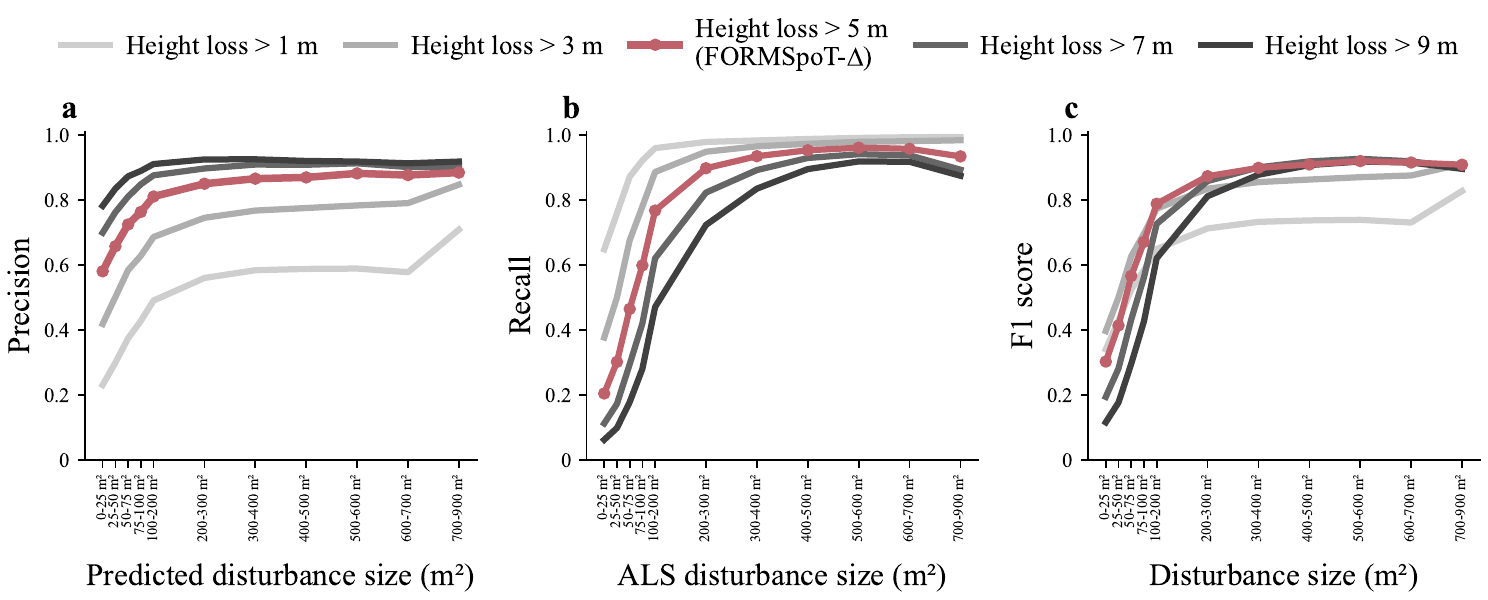}
    \caption{\textbf{Disturbance detection performance with varying height-loss thresholds.} Precision (a), recall (b), and F1-score (c) computed using a 30~m grid-based evaluation against ALS reference disturbances (\cref{sec:methods:grid_based_tc_change}) across all 19 test sites. Each line corresponds to a height-loss threshold from 1~m to 9~m. The threshold of 5~m, used to compute \fdelta polygons, is highlighted in red.}
    \label{fig:precision_recall_f1_tc_change_tresholds}
\end{figure*}

\cref{fig:precision_recall_f1_tc_change_tresholds} shows precision, recall, and F1-score as a function of height-loss threshold and disturbance size class across all sites. The 5~m threshold maximizes F1-score across all size classes, confirming its suitability for \fdelta. Lower thresholds increase recall but reduce precision, reflecting sensitivity to residual acquisition artifacts; higher thresholds improve precision at the cost of missing moderate height-loss events.

Detection performance increases with disturbance size, with a marked transition around a size of 100\mtwo. Below this, events in the 0-25\mtwo class show moderate precision (0.58) and low recall (0.21), reflecting difficulty in separating small height losses from acquisition noise. Performance improves considerably within the sub-100\mtwo range, reaching a precision of 0.76 and recall of 0.60 for the 75-100\mtwo class. Above 100\mtwo, the F1 score is above 0.8 with a precision that remains relatively stable while recall continues to increase, indicating that false detections are well controlled at larger scales and that the main remaining limitation is sensitivity to smaller events.

The following section examines how our high resolution polygons compares to existing coarser-resolution satellite products, and quantifies the fraction of disturbance activity structurally inaccessible to them.

\subsection{Comparison with coarser resolution products}
\subsubsection{ALS-based disturbance validation}
\label{sec:results:als_validation_disturbance}

\begin{figure*}[!t]
  \centering
  \includegraphics[width=\linewidth]{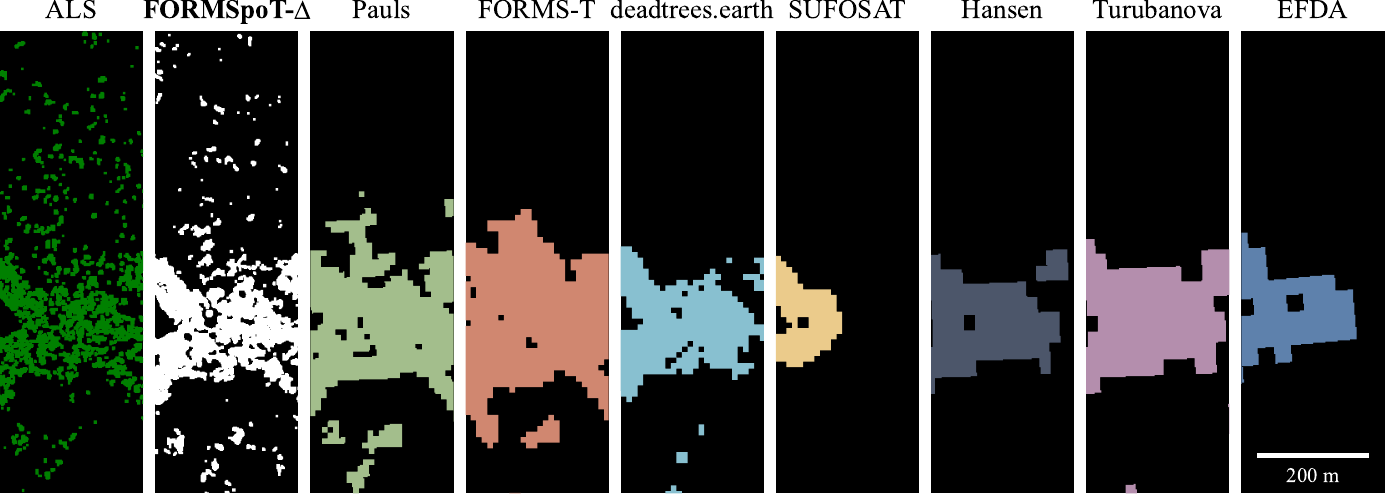}
  \caption{\textbf{Visual comparison of disturbance polygons at the Joux-Fresse site (2019-2022).} ALS-derived polygons were derived using a 5~m threshold and an opening morphological filter (\cref{sec:methods:grid_based_tc_change}). The Turubanova dataset is only available for 2019-2021 but is shown for reference. Location: 46.83 \textdegree{}N, 6.03\textdegree{}E.}
  \label{fig:6-visual_comparison_sota}
\end{figure*}

\begin{figure*}[!t]
  \centering
  \includegraphics[width=\linewidth]{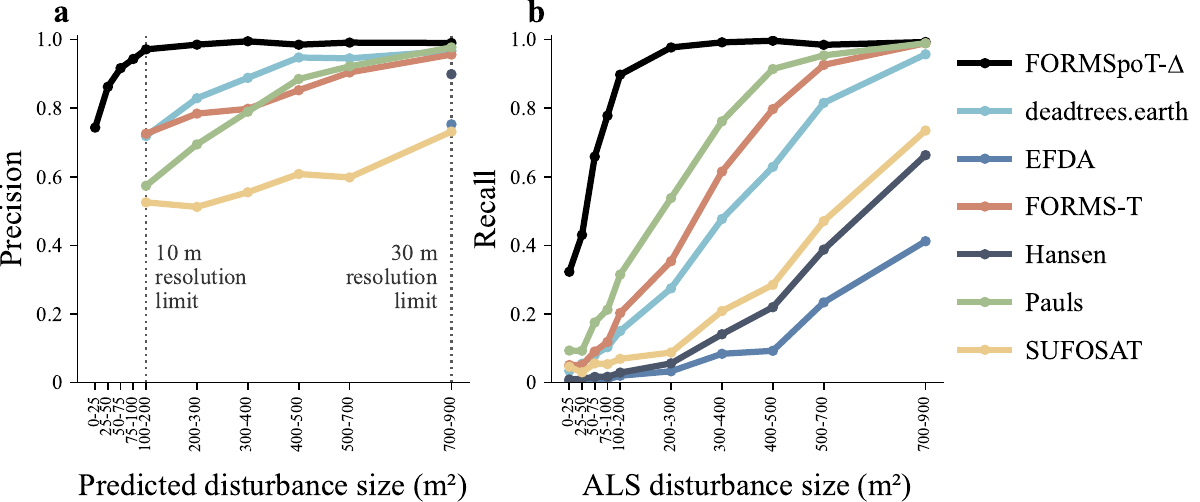}
  \caption{\textbf{Detection performance across disturbance size classes at the Joux-Fresse site.} Precision (a) and recall (b) as a function of disturbance size class for \fdelta and six existing products, computed on a 30~m grid (\cref{sec:methods:grid_based_tc_change}). Vertical dotted lines indicate the theoretical minimum detectable area at 10~m and 30~m resolution.}
  \label{fig:7-recall_precision_joux_fresse}
\end{figure*}

\cref{fig:6-visual_comparison_sota} illustrates disturbance polygons from \fdelta and six existing products alongside ALS-derived reference disturbances at the Joux-Fresse site. \fdelta closely captures the spatial pattern of ALS observed disturbances, including numerous small disturbance events. Other products at 10~m resolution (Pauls, \formst, deadtrees.earth) detect the largest disturbances but miss most smaller events, while Landsat-based products (Hansen, EFDA, Turubanova) are limited to disturbances exceeding roughly 0.1~ha. 
\cref{fig:7-recall_precision_joux_fresse} shows precision and recall as a function of disturbance size class at the Joux-Fresse site. For every product, the disturbance size at which detection becomes reliable is much larger than the native resolution of the sensor it is derived from. For 10~m resolution products, recall remains low for the 100-200\mtwo class (0.20 for \formst, 0.31 for Pauls, 0.15 for deadtrees.earth, 0.06 for SUFOSAT) and only reaches high levels in the 500-700\mtwo range (e.g., 0.95 for Pauls), suggesting that reliable detection requires multiple connected disturbed pixels. Landsat-based products show some detection below their native resolution of 30 m (recall~=~0.39 for the 500-700\mtwo class) but still miss a large fraction of disturbances at their native resolution (recall~=~0.66 for Hansen above 700\mtwo). In contrast, \fdelta reaches a recall of 0.90 in the 100-200\mtwo class. Precision is high across all size classes above 100\mtwo (0.97 for the 100-200\mtwo class) and remains high even for smaller events (0.74 for the 0-25\mtwo class). Coarser products show moderate precision below 500\mtwo, improving only at larger sizes; Landsat-based products achieve high precision at their native resolution (Hansen: 0.90 above 900\mtwo). Together, these results confirm that the practical detection frontier lies substantially above the nominal pixel-size limit for all products.

\begin{figure*}[t]
  \centering
  \includegraphics[width=\textwidth]{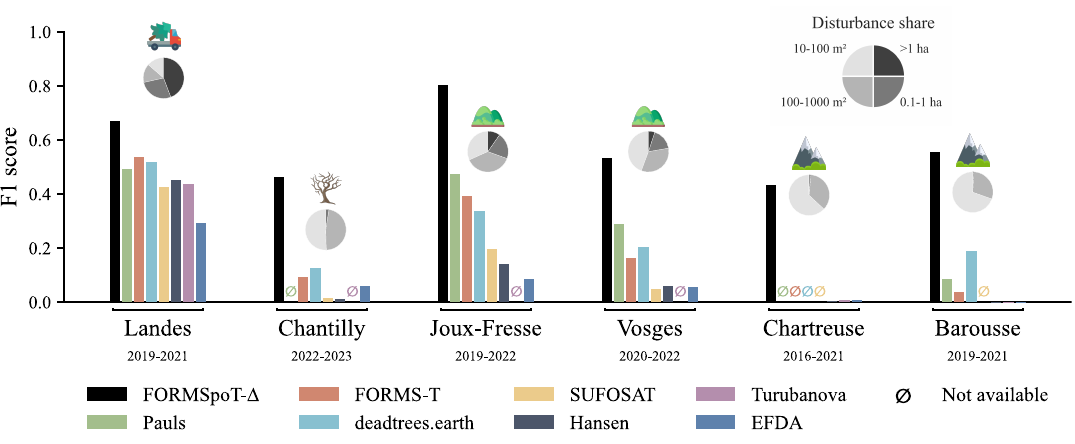}
  \caption{\textbf{F1-scores across the representative test sites.} Datasets that could not be evaluated at a given site due to mismatched acquisition dates are marked with $\varnothing$. For each site, the proportion of disturbance size classes as a percentage of total disturbed area derived from ALS polygons is shown above the plot. From left to right, symbols correspond to a heavily managed maritime pine plantation, a declining oak forest with salvage logging, a low-mountain mixed forest, and a high-mountain mixed forest with selective logging.}
  \label{fig:8-F1_scores}
\end{figure*}
These patterns are consistent across the six highlighted test sites (\cref{fig:8-F1_scores}; see Supplementary Figure~S4 for the remaining sites), though the relative importance of each size class varies with forest type and terrain. At the Landes site, dominated by large clearcuts (disturbances~\textgreater~0.1~ha accounting for 72\% of total disturbed area), F1-scores are relatively high across most products (e.g. 0.49 for Pauls, 0.42 for SUFOSAT, 0.45 for Hansen), with \fdelta reaching a F1 of 0.67. Performance diverges at structurally complex sites: at Chantilly, where disturbances consist mainly of salvage logging after individual tree mortality, only \fdelta achieves a significant F1-score (0.46), as the fine-scale signature of these events falls below the detection threshold of coarser products. The contrast is most pronounced at high-mountain sites such as Barousse, where 69.5\% of disturbed area occurs in polygons smaller than 100\mtwo: 10~m and 30~m products achieve low F1-scores (best was deadtrees.earth at 0.18) while \fdelta reaches 0.56. This site-level variation confirms that the performance gap between sensor resolutions scales with the proportion of fine-scale disturbances.

\subsubsection{Sensitivity to low-intensity disturbances: validation against NFI plot revisits}
\label{sec:NFI_validation_polygons}

\begin{figure}[!t]
  \centering
  \includegraphics[width=0.9\linewidth]{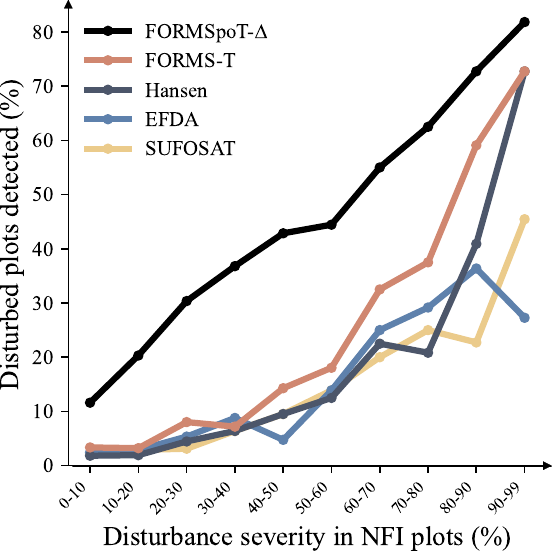}
  \caption{\textbf{Detection of low-intensity disturbances from French NFI plot revisits (2018-2023).} Fraction of disturbed plots detected as a function of disturbance severity for \fdelta and existing products. Disturbance severity is defined as the percentage of trees recorded as cut or fallen over the five-year revisit interval. Each point represents the fraction of NFI plots within that severity class whose center intersects a predicted disturbance polygon.}
  \label{fig:9-NFI_disturbances}
\end{figure}

\fdelta achieves an overall F1-score of 0.44 against NFI-recorded disturbances over 2018-2023 (precision: 0.67, recall: 0.33). Detection rates increase consistently with disturbance severity, defined as the percentage of trees removed in the plot (\cref{fig:9-NFI_disturbances}): \fdelta detects 73\% of plots where 80-90\% of trees were removed, compared to only 11\% where fewer than 10\% were lost. This gradient is expected, as the removal of a single tree does not necessarily produce a canopy gap detectable with a 5~m height loss threshold, depending on tree size and position relative to other trees. The positional uncertainty in the NFI plot locations can also degrade the detection of partial harvests and scattered canopy openings. Nonetheless, \fdelta maintains higher detection rates than all other products across every intensity class: it detects 20\% of plots with only 10-20\% tree loss, compared to at most 3\% for other products, reflecting the advantages of HR imagery.

\subsection{Spatio-temporal patterns of forest disturbances in France}
\label{sec:results:spatio_temporal_patterns}

\subsubsection{National-scale disturbance size regime}
\begin{figure}[h]
    \centering
    \includegraphics[width=\linewidth]{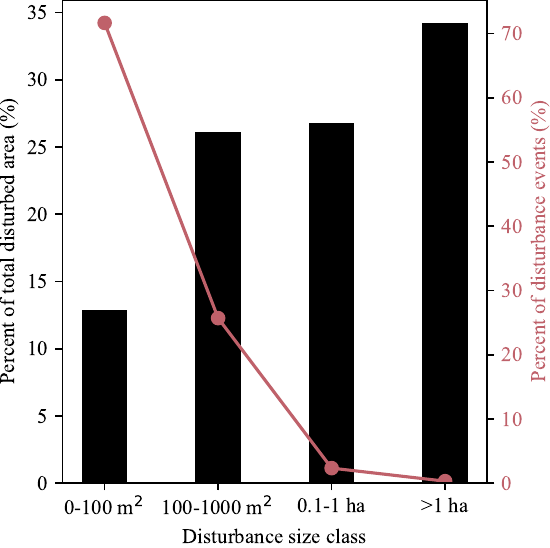}
    \caption{\textbf{Small disturbances dominate in number but not in area across French forests.} Percent of disturbed area (black bars) and percent of disturbance events (red line) per size class, derived from \fdelta, averaged over 2015-2023. The first and last years of the time series (2014 and 2024) were excluded due to edge effects of the TV denoising process.}
    \label{fig:results:dist_count_and_area_per_class}
\end{figure}

Across mainland France, \fdelta records an average of 5 million disturbance events yearly, spanning four orders of magnitude in area (\cref{fig:results:dist_count_and_area_per_class}). Small disturbances dominate numerically: events below 100\mtwo account for 72\% of all polygons but only 13\% of total disturbed area. Disturbances below 1,000\mtwo (0.1 ha), detected with low reliability at 30~m resolution, represent 97\% of the events and 39\% of the total area. Large disturbances, exceeding 10,000\mtwo (1~ha), represent 0.3\% of events and contribute 34\% of total disturbed area. Operational disturbance products based on Landsat 30 m resolution imagery (Hansen and EFDA), therefore miss the vast majority of disturbance events and almost half of the disturbed area, which occur at scales below the size of their pixels.

\begin{figure*}[p]
    \centering
    \includegraphics[width=\linewidth]{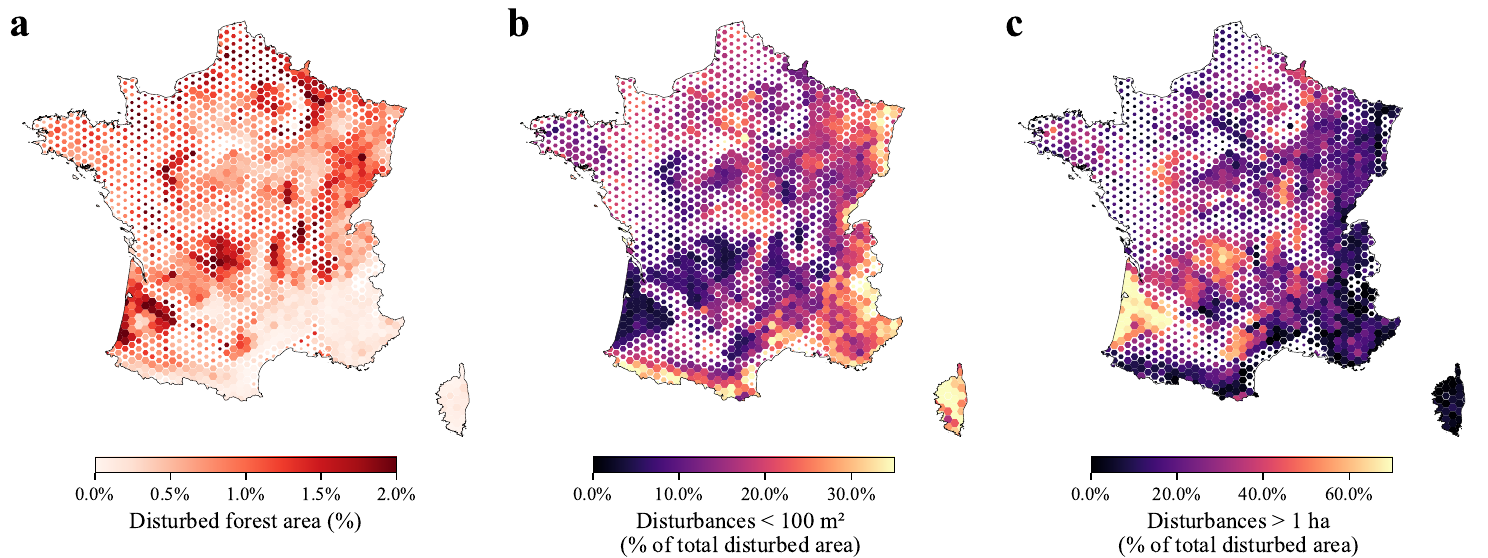}
    \caption{\textbf{Spatial structure of forest disturbance regimes across France (2015-2023).} (a) Mean annual fraction of forest area disturbed per hexagon. (b) Share of small disturbances (<100\mtwo) in total disturbed area per hexagon. (c) Share of large disturbances (>1~ha) in total disturbed area per hexagon. All maps show mean values over the 2015-2023 period. The first and last years of the time series (2014 and 2024) are excluded due to edge effects of the TV denoising process. Hexagon size scales with forest fraction; hexagons are shown at full size when forest fraction exceeds 40\%.}
    \label{fig:results_forest_management_static}
\end{figure*}

The mean annual fraction of forest area disturbed varies strongly across France (\cref{fig:results_forest_management_static}a). The Mediterranean basin and Corsica show the lowest disturbance rate, while the Landes (southwest), Normandy (northwest), and parts of northern France show the highest, up to 2\% yearly. The spatial distribution of disturbance size classes is also markedly heterogeneous (\cref{fig:results_forest_management_static}b,c). Small disturbances ($<$100\mtwo) constitute the dominant share of disturbed area in mountain regions, particularly in the Alps, Pyrenees, and Massif Central (\cref{fig:results_forest_management_static}b). Conversely, large disturbances ($>$1~ha) dominate in the Landes and, to a lesser extent, in the Limousin and Périgord regions (\cref{fig:results_forest_management_static}c). The full breakdown of disturbance size class fractions across all regions and a comparison of disturbance regime with a 30~m based disturbance detection system \citep{hansenHighResolutionGlobalMaps2013} is provided in \ref{sec:appendix:disturbance_regimes}.

\subsubsection{Temporal dynamics and the bark beetle crisis}

\begin{figure*}[p]
    \centering
    \includegraphics[width=\linewidth]{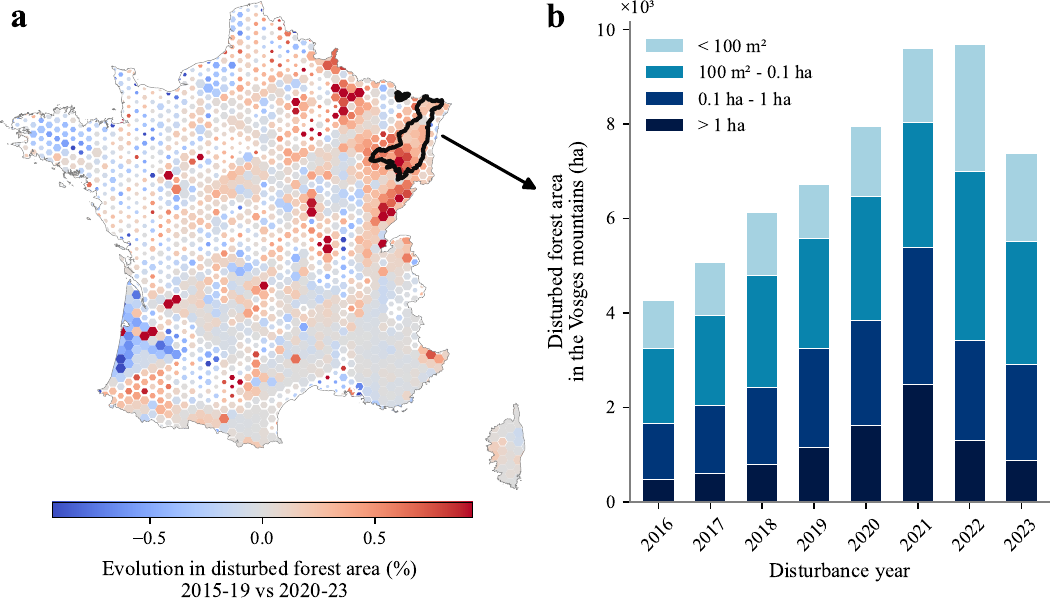}
    \caption{\textbf{Regional intensification of disturbances and the bark beetle crisis in the Vosges.} (a) Change in mean annual disturbed forest area fraction between the 2015-2019 and 2020-2023 periods. The first and last years of the time series (2014 and 2024) are excluded due to edge effects of the TV denoising process. Red hexagons indicate an increase in disturbance activity. Hexagon size scales with forest fraction; hexagons are shown at full size when forest fraction exceeds 40\%. The Vosges region boundary is overlaid. (b) Annual disturbed area in the Vosges by disturbance size class, showing the temporal signature of tree height loss (salvage logging) after bark beetle outbreak.}
    \label{fig:results_multitemporal_vosges}
\end{figure*}

Disturbance activity increased substantially between the 2015-2019 and 2020-2023 periods in northeastern France, with the strongest signal in the Vosges and Argonne forest regions in northeastern France where a bark beetle crisis happened in 2017-2022 (\cref{fig:results_multitemporal_vosges}a). The Landes (southwest) shows the opposite trend, with a decrease in disturbed area over the same period.
The Vosges time series indicates a pronounced disturbance peak in 2021 and 2022 (\cref{fig:results_multitemporal_vosges}b). Total disturbed area in 2022 reached 9,680~ha, compared to 4,259~ha in 2016, representing a 2.3-fold increase. This peak is driven by all disturbance classes, but small and medium disturbances (< 0.1 ha) dominate the total disturbed area in 2022 (65\% of the disturbances). The regional temporal dynamics for all regions are provided in Supplementary Section~6.

\section{Discussion}
\subsection{The resolution frontier: what fraction of French forest disturbances remains invisible from space with current 10-30 m satellite monitoring products?}

The national-scale size distribution (\cref{sec:results:spatio_temporal_patterns}) shows that disturbances below 1,000\mtwo (0.1~ha), i.e. structurally below the pixel size of Landsat-based products (30~m), represent the vast majority of events (97\%) and a significant proportion (39\%) of total disturbed area in French forests. Within this sub-pixel range, the finest events below 100\mtwo still account for the majority (72\%) of all polygons. Current products at 10-30~m resolution therefore poorly detect most disturbance events and a substantial fraction of the disturbed area.
This limitation appears to be predominantly sensor-driven rather than algorithmic: despite different architectures and training strategies, \formst, Pauls, and deadtrees.earth converge toward the same effective detection threshold near 500-700\mtwo at the Joux-Fresse site (\cref{fig:7-recall_precision_joux_fresse}). This convergence of independently designed products toward a common limit suggests that the binding constraint is the information content of Sentinel-1/2 10~m imagery rather than any single algorithmic choice. Algorithm design, minimum mapping units, and post-processing also shape the effective detection threshold of each product, but they are unlikely to explain why methods with very different architectures plateau at the same scale. This result is consistent with the current literature: \citet{mercierMappingSmallsizedLogging2026} found that a Sentinel-1 framework reached 90\% detection only at 0.20~ha (twenty times the pixel area) and most disturbance products impose a minimum mapping unit of several pixels to control false detection that could occur with a single pixel \citep{reicheForestDisturbanceAlerts2021, senfMappingForestDisturbance2020, viana-sotoEuropeanForestDisturbance2025}.
This detection limit is not uniformly distributed across forest types. It is most pronounced in mountains, where the dominant scale of canopy openings is small for two converging reasons. On the natural side, gap-forming processes such as single-tree mortality and small windthrows shape much of the disturbance regime in mountain forests \citep{bebiChangesForestCover2017}, with reported small patch size in subalpine pine stands \citep{rathgeberSpatiotemporalGrowthDynamics2003} and in protected Alpine forests \citep{maroschekQuantifyingPatchSize2024}. On the management side, steep terrain constrains mechanized harvesting and favors lower-intensity, spatially constrained extraction \citep{leversDriversForestHarvesting2014}, while uneven-aged silviculture based on selection cutting has become a widespread practice in European mountain forests \citep{redonGeneralisationUnevenagedManagement2014}. The relative contribution of natural and managed processes to the small-patch signal cannot be disentangled from height-loss observations alone, but both push the disturbance size distribution well below the resolution of current 10-30~m products (\cref{fig:results_forest_management_static}.b). Consistent with this, at the Barousse (Pyrenees) and Chartreuse (Alps) sites, where the majority of disturbed area occurs in sub-100\mtwo events, coarser products achieve F1-scores below 0.20 (\cref{fig:8-F1_scores}). These small-scale disturbances hold an important ecological role \citep{muscoloReviewRolesForest2014} and dominate in European unmanaged forests \citep{hobiGapPatternLargest2015}, yet they have been largely absent from country- or continental-scale satellite monitoring records \citep{korznikovPitfallsForestDamage2024, yamadaCausalAnalysisAccuracy2020}.
\fdelta extends reliable detection down to approximately 100\mtwo, making observable at the national scale a disturbance size class that accounts for the majority of events in French forests. However, the threshold sensitivity analysis (\cref{fig:precision_recall_f1_tc_change_tresholds}) and the visual interpretation (\cref{fig:results:viz_fdelta-ljlf_barousse}.b) show that the finest events, individual mortality for smaller trees in dense canopies, remain at the edge of what height-loss detection from passive optical imagery can recover at this resolution.

\subsection{Reliability and limits of annual HR disturbance monitoring}


The ALS grid-based (\cref{fig:precision_recall_f1_tc_change_tresholds}) and the NFI (\cref{fig:9-NFI_disturbances}) evaluations reveal a sharp transition in detection reliability for events smaller than 100\mtwo. Below this threshold, there is an increasing difficulty of separating genuine height losses from residual acquisition noise as disturbance size approaches the pixel scale. This could be explained by the following points: 

First, the systematic negative bias observed for tall canopies (-9.8~m in the 35-40~m reference height class, \cref{fig:5-NFI_static_validation}b), is a known saturation effect of passive optical sensors for tall canopies, even combined with deep learning algorithms \citep{schwartzRetrievingYearlyForest2025, wagnerWalltowallAmazonForest2025, langCountrywideHighresolutionVegetation2019, wanSatellitebasedMappingAnnual2025}. Because \fdelta relies on a fixed 5~m height-loss threshold, this saturation effect can cause real disturbances to fall below the detection threshold in mature stands with canopy heights above 25~m.

Second, tree cover is systematically overestimated in dense canopies (\cref{tab:results:tc_validation}). This indicates that small within-canopy gaps are not resolved at 1.5~m resolution, making partial crown removal effectively invisible in already closed canopies.

Third, acquisition heterogeneity introduces noise that propagates into the height predictions. The annual \spotsixseven mosaics are assembled from images acquired across different months, producing phenological mismatches between years. Similarly, variations in sun angles, atmospheric conditions, and viewing geometry affect shadow patterns and reflectance, while specific cases of cloud or snow contamination further degrade local predictions, as observed over western Corsica in 2024. 

Fourth, the ALS reference is not free of uncertainty. Resampling the ALS CHMs from 1~m to 1.5~m by taking the maximum value reduces the detection of small reference gaps. Moreover, the 19 sites span a wide range of acquisition configurations, and changes between two successive ALS acquisitions could introduce apparent canopy differences unrelated to real disturbances. These effects can be amplified in mountainous terrain, where slope-induced occlusion further degrades small-gap detection. 

Fifth, \fdelta defines disturbance as a measurable canopy height loss, and therefore captures the moment when trees are physically removed from the canopy rather than the moment of tree death itself. This distinction is particularly relevant for biotic disturbances such as bark beetle outbreaks: infested trees retain their height for months to years after death and only contribute to a height-loss signal once they are extracted through salvage logging, collapse, or lose their crown. In managed forests, the time lag between tree death and detection is therefore largely controlled by the speed of salvage operations, while in protected areas where infested trees are left standing, the detection may be delayed by several years. The temporal signature observed in \fdelta during the 2017-2022 bark beetle crisis in northeastern France (\cref{fig:results_multitemporal_vosges}) thus reflects the timing of wood mobilization and stand structural breakdown rather than the underlying outbreak dynamics. The same caveat applies, to a lesser extent, to drought-induced mortality and other slow forms of canopy decline.

Finally, the TV denoising step involves an inherent trade-off. The temporal regularization ($\lambda_{\text{temp}} = 5$) effectively suppresses acquisition-driven noise, as reflected in the low interannual variability of height errors (MAE SD = 0.08~m across years, \cref{fig:5-NFI_static_validation}c). However, this same regularization can attenuate real abrupt changes, particularly at the edges of the time series (2014 and 2024), where the algorithm has fewer neighboring years to distinguish signal from noise. 

More fundamentally, increasing spatial resolution changes the nature of the signal itself. At 1.5~m, reflectance is less spatially aggregated and therefore more sensitive to local structural variability, illumination, and acquisition geometry than at coarser scales, where averaging smooths noise and stabilizes distributions \citep{woodcockFactorScaleRemote1987}. This makes per-pixel height prediction intrinsically more variable at 1.5~m than at 10-30~m, and sets a floor on the precision achievable for the finest disturbance events regardless of model architecture (see also \citep{besicPolarimetricIncoherentTarget2015} for an analogous effect in SAR).

\subsection{Disturbance regimes across French forests}

Across mainland France, disturbance regimes vary sharply with forest type and management context (\cref{fig:results_forest_management_static}). At one end of the spectrum, the Mediterranean basin and Corsica show low disturbance rates (\textless 0.1\% yr\textsuperscript{-1}) consistent with the long-term land-use abandonment and absence of forest management that has shaped these landscapes since the mid-19th century \citep{sanromansanzEcologySocietyLongTerm2013}. At the other extreme, the Landes (southwest) shows the highest disturbance rates (\textgreater~2\% yr\textsuperscript{-1}), which could be linked with clearcuts in the largest plantation forest in Europe, where maritime pine is cultivated in even-aged monocultures under short rotations \citep{petuccoLandExpectationValue2018a, ansaldiOptimisationForestManagement2025}. In the Limousin and Périgord, large disturbance events could reflect the maturation of post-war spruce plantations \citep{onfGrandeHistoireForets2022a} and chestnut coppice rotations. Mountain forests present a fundamentally different regime:
sub-100\mtwo events contribute more strongly to the disturbed area in the Alps, Pyrenees, Vosges, and Jura. It could be explained with terrain constraints for wood exploitation \citep{dincaTimberHarvestingMountainous2025} which leave space for scattered natural mortality events resulting in small canopy gaps formation and selective logging practices.
Because these fine-scale events are poorly detected with 10-30~m products, disturbance monitoring in these regions has been nearly absent from existing country-scale satellite products (\cref{sec:results:als_validation_disturbance}).

Continental and global-scale Landsat-based products \citep{senfMappingForestDisturbance2020, viana-sotoEuropeanForestDisturbance2025, hansenHighResolutionGlobalMaps2013} qualitatively agree with these patterns: \citet{hansenHighResolutionGlobalMaps2013} identifies the Landes, the Limousin-Périgord area, and northeastern France as the most disturbed regions (\cref{fig:appendix_hanscen_comparison}) but two differences emerge. In mountain regions, \fdelta shows substantial disturbance activity that is poorly detected with 30~m products, as these sub-100\mtwo events fall below the detection threshold documented in \cref{fig:7-recall_precision_joux_fresse}. Conversely, in the Landes, \fdelta estimates less total disturbed area than Hansen, likely because 30~m pixels represent clearcut boundaries as coarse polygons that overestimate the actual disturbed footprint, whereas 1.5~m polygons delineate harvest edges more tightly (\cref{fig:appendix_hanscen_comparison}). This underestimation effect has already been observed by \citet{vanderwoudeEuropeanForestDisturbance2026a} when comparing their 10 m Sentinel-1 product with Hansen and EFDA.

Anthropogenic and natural forest disturbances in Europe have increased over the past decades \citep{senfMappingForestDisturbance2020, pataccaSignificantIncreaseNatural2023} and are projected to more than double by the end of the century under climate change \citep{grunigClimateChangeWill2026}. Our 11-year time series is too short to establish long-term trends, but it captures the temporal signature of recent disturbance events such as the bark beetle crisis in northeastern France in the 2017-2022 period \citep{arthurSpatialRemoteSensing2024}. The disturbance peak detected in \fdelta likely corresponds to the wave of sanitary logging that followed the outbreak rather than to tree death itself (\cref{fig:results_multitemporal_vosges}), since standing dead trees only contribute to a height-loss signal once removed or collapsed.

Additionally, the 1.5 m resolution can observe dynamics that remain hidden in coarser products. In the Vosges, the 2022 disturbance peak is not driven by an increase in large harvest events, which remain relatively stable, but by a surge in small and medium-sized disturbances (<0.1~ha) that account for 65\% of the total disturbed area in 2022 (\cref{fig:results_multitemporal_vosges}.b). More broadly, background mortality and fine-scale gap expansion generate disturbances that fall below the detection capabilities of existing satellite products. By making these events visible at the national scale, \fdelta offers a way to track whether the anticipated intensification of disturbances under climate change \citep{grunigClimateChangeWill2026} manifests primarily through more frequent large events or through a diffuse increase in fine-scale mortality, two scenarios with very different ecological and management implications.

\subsection{Implications for forest monitoring and carbon accounting}

Sub-100\mtwo disturbances account for 13\% of total disturbed area nationally (\cref{fig:results:dist_count_and_area_per_class}), and this fraction is likely a lower bound given that \fdelta itself underdetects the finest events (\cref{fig:precision_recall_f1_tc_change_tresholds}). In tropical forests, \citet{espirito-santoSizeFrequencyNatural2014} showed that, in the absence of management, small canopy gaps dominate carbon losses. In temperate managed forests, this contribution is smaller but remains absent from current continental-scale satellite-based estimates. EU carbon reporting relies on a combination of NFI extrapolations and satellite-derived disturbance products \citep{blujdeaEUGreenhouseGas2015, korosuoRoleForestsEU2023}, both of which have limited sensitivity to fine-scale events: NFI estimates are designed to be statistically robust only when aggregated to regional or national scale \citep{bontempsTakeFiveBeat2024}, while 10-30~m products miss the prevailing disturbance regime in mountain forests. By providing wall-to-wall annual disturbance maps at a fine scale, country-wide HR approaches like FORMSpoT could complement NFI observations to improve both the spatial precision and temporal resolution of carbon flux estimates in these regions.

Beyond carbon, \fdelta could contribute to a spatial indicator of forest management practices. Higher resolution enables the identification of thinning, selective logging, and patterns associated with coppice systems \citep{evansCoppiceForestryOverview1992} or even-aged and uneven-aged stands \citep{noletComparingEffectsEven2018}. Such information is important to adapt forest practices to climate change, as thinning is thought to influence forest resilience \citep{moreauOpportunitiesLimitationsThinning2022}. In France, where 75\% of forests are privately owned in small, fragmented parcels, spatially explicit disturbance data could help public authorities better understand management patterns and design targeted policies for climate adaptation. \fdelta can also extend ALS-based gap dynamics studies \citep{krugerGapExpansionDominant2024} to the national scale at lower cost. However, height change alone cannot attribute disturbances to specific causes. Combining \fdelta with existing species maps such as the French BD Forêt \citep{ignBDForetVersion2018} and climate anomaly indices would enable a transition from disturbance detection toward disturbance attribution, which links observed canopy losses to their ecological or anthropogenic drivers.

\section{Conclusion}

In this work, we presented a workflow to generate high-resolution canopy height maps (FORMSpoT) and disturbance polygons (\fdelta) across France from eleven years of \spotsixseven imagery.
We first showed that existing 10-30~m products reach reliable disturbance detection only above 500-900\mtwo, leaving a substantial fraction of forest disturbance activity undetected, particularly in mountain regions where sub-100\mtwo events dominate. 
Second, we demonstrated that a national-scale HR time series can provide reliable annual disturbance monitoring for events above approximately 100\mtwo, although detection degrades rapidly below this threshold. Third, aside from natural disturbances, results are consistent with dominant management strategies found in France: clear-cut-dominated dynamics in large plantations like the Landes and diffuse, fine-scale mortality or selective logging in mountain forests that was previously detected with low reliability with satellite-based monitoring.
Both FORMSpoT and \fdelta are openly available at \url{https://doi.theia.data-terra.org/FORMSpoT/}, making them accessible to a wide range of forestry stakeholders. 

\section{Acknowledgements}

We acknowledge all contributors to this work. Access to the Chantilly ALS acquisition was provided by Laurent Saint-André and the group Ensemble sauvons la forêt de Chantilly. ALS data were also provided by the Office National des Forêts (ONF), with coordination by Jérôme Bock. The Massif des Bauges lidar dataset was funded by ADEME through the GRAINE program (PROTEST project 1703C0069), the Jura-Doubs dataset by the Conseil Régional Bourgogne Franche-Comté, and the Chartreuse dataset by ONF, PNR de Chartreuse, and INRAE. We thank DINAMIS and Airbus for open access to \spotsixseven imagery, the French National Forest Inventory foresters for field data collection, and IGN for nationwide access to LiDAR HD data. This work is part of the ALAMOD project (FairCarboN) funded by ANR under France 2030 (ANR-22-PEXF-0002), with additional support from the PC MONITOR of the PEPR FORESTT (ANR-24-PEFO-0003), AI4FOREST (ANR-22-FAI1-0002-01), TOSCA CFOREST 50M, SHARP (PEPR IA, ANR-23-PEIA-0008), the ARTEMIS program of Lorraine Université d’Excellence (ANR-15-IDEX-04-LUE), Hi! PARIS (ANR-23-IACL-0005), the EU Horizon Europe program (Grant No. 101131841), and the Swiss State Secretariat for Education, Research and Innovation (SERI). Computational resources were provided by IDRIS under the HPC allocations 2024-AD010114718 and 2025-AD010114718R1 granted by GENCI. Finally, we thank Flaticon.com for the icons used in \cref{fig:8-F1_scores} and \cref{tab:1_ALS-sites}.

\

\appendix
\onecolumn
\section{Metrics used for the grid-based validation of disturbance polygons}
\label{sec:appendix:metrics_f1_precision_recall_gridbased}
Each 30~m pixel is first labeled as disturbed or undisturbed. A pixel is considered disturbed whenever its disturbed area fraction is strictly positive, i.e. as soon as any disturbance is detected within it. This applies identically to the ALS reference and to every evaluated product. We denote by $ALS_{dist}$ the set of disturbed reference pixels and by $F\Delta_{dist}$ the set of disturbed pixels from \fdelta (or from any other evaluated product). $|\cdot|$ denotes the number of pixels in a set.

Each disturbed pixel is then assigned to a size class (e.g. 100-200\mtwo) from its disturbed area fraction. Because precision and recall measure complementary quantities, they rely on a different size reference. Recall asks, among reference disturbances of a given size, how many are retrieved, so size is taken from the ALS reference fraction, giving the subset $ALS_{dist_{class}}$. Precision asks, among the product detections of a given size, how many are correct, so size is taken from the evaluated product fraction, giving the subset $F\Delta_{dist_{class}}$. A consequence is that 30~m products (e.g. Hansen, EFDA) only have a single, whole-pixel size class for precision.

Precision ($P_{class}$), recall ($R_{class}$), and F1-score ($F1_{class}$) per size class are then defined as:

\begin{align}
    P_{class} &= \frac{|ALS_{dist} \cap F\Delta_{dist_{class}}|}{|F\Delta_{dist_{class}}|} \\
    R_{class} &=\frac{|ALS_{dist_{class}} \cap F\Delta_{dist}|}{|ALS_{dist_{class}}|} \\
    F1_{class} &=\frac{2\times P_{class} \times R_{class}}{P_{class}+R_{class}}
\end{align}

The same metrics were applied to all other disturbance datasets to evaluate their ability to detect sub-pixel disturbances.

\section{Effect of TV Denoising on height predictions}
\label{sec:appendix_TV_effects}
\begin{figure}[H]
  \centering
  \includegraphics[width=\textwidth]{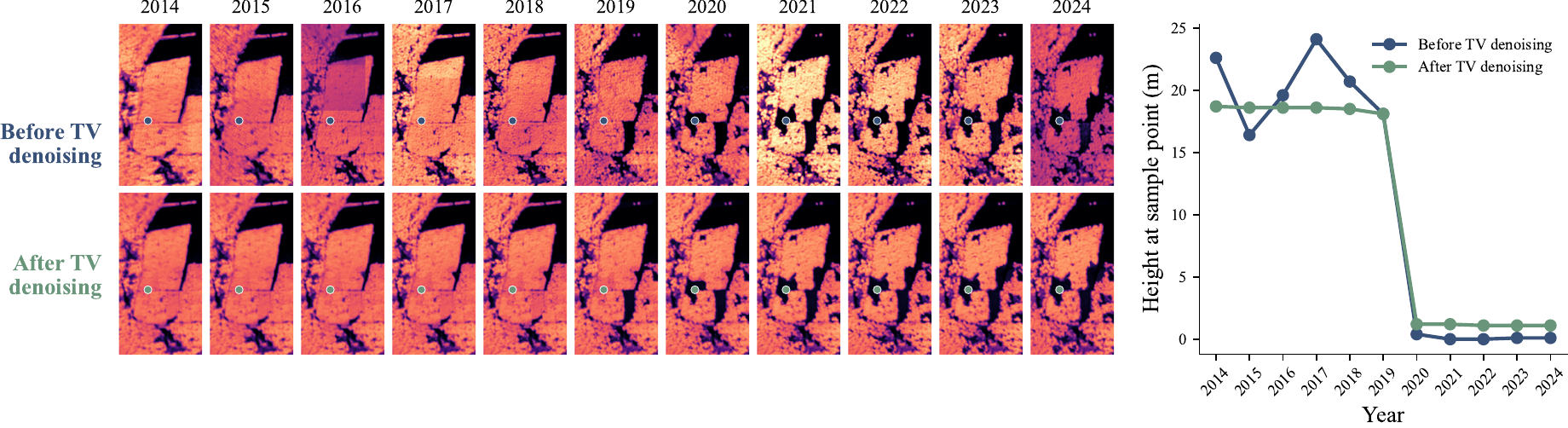}
  \caption{Visual comparison of our height predictions before and after applying the TV denoising algorithm used as post-processing for the timeseries (location: 48.50 N, 7.14 E). The right panel shows the evolution of the height sampled at a given location.}
  \label{fig:11-TV_effect}
\end{figure}

To ensure spatial and temporal consistency in the FORMSpoT height time series, we applied a spatio-temporal denoising algorithm (see \cref{sec:TV_denoising}). \cref{fig:11-TV_effect} illustrates its effect on a specific location: before denoising, the series shows large fluctuations caused by changing \spotsixseven acquisition conditions, while after denoising, the height becomes stable and coherent through time. This consistency is essential for downstream analyses such as disturbance detection based on abrupt height changes.

\newpage
\section{Tree cover validation}
\label{sec:appendix:tc_validation}
Across all sites, FORMSpoT 30~m aggregated tree cover estimates agree well with ALS-derived tree cover (MAE~=~0.09, r\textsuperscript{2}~=~0.83, see supplementary Figure~S3). 
Agreement is consistent across cover classes with a MAE ranging from 0.03 (0.9-1 class) to 0.245 (0.5-0.6 class). Though, tree cover is systematically overestimated at high values. 
For instance, 71.2\% of pixels with ALS tree cover in the 0.7-0.8 range are predicted in the 0.9-1 class by FORMSpoT. It implies that smaller canopy gaps visible in ALS CHMs, cannot be resolved by the 1.5 m FORMSpoT resolution with the 5~m height threshold applied to compute tree cover. As a result, \fdelta is likely to underdetect very small disturbances in already dense canopies, where partial crown removal may not produce a sufficient gap signature (See \cref{sec:results:fdelta}) 
\begin{table}[h]
    \centering
    \caption{\textbf{Validation of FORMSpoT tree cover with ALS data in 30 m pixels.} Confusion matrix comparing FORMSpoT and ALS-derived tree cover at 30~m resolution across all test sites. Each cell $(i,j)$ gives the percentage of pixels in ALS tree cover class $j$ assigned to FORMSpoT tree cover class $i$. Values on the diagonal indicate correct classification; off-diagonal values indicate over- or underestimation.}
{\footnotesize\setlength{\tabcolsep}{3.5pt}%
\begin{tabular}{@{}cccccccccccc@{}}
\multicolumn{2}{c}{} & \multicolumn{10}{c}{ALS tree cover} \\
\cmidrule(lr){3-12}
& & 0.0--0.1 & 0.1--0.2 & 0.2--0.3 & 0.3--0.4 & 0.4--0.5 & 0.5--0.6 & 0.6--0.7 & 0.7--0.8 & 0.8--0.9 & 0.9--1.0 \\
\toprule
\multirow{10}{*}{\rotatebox{90}{FORMSpoT tree cover}} & 0.0--0.1 & \cellcolor[HTML]{208A52}\textcolor{white}{92.0} & \cellcolor[HTML]{B1D6C3}\textcolor{black}{32.3} & \cellcolor[HTML]{DCEDE4}\textcolor{black}{14.5} & \cellcolor[HTML]{EDF6F1}\textcolor{black}{7.6} & \cellcolor[HTML]{F6FAF8}\textcolor{black}{4.2} & \cellcolor[HTML]{FAFDFB}\textcolor{black}{2.2} & \cellcolor[HTML]{FDFEFE}\textcolor{black}{1.1} & \cellcolor[HTML]{FDFEFE}\textcolor{black}{0.8} & \cellcolor[HTML]{FFFFFF}\textcolor{black}{0.3} & \cellcolor[HTML]{FFFFFF}\textcolor{black}{0.2} \\
& 0.1--0.2 & \cellcolor[HTML]{F8FCFA}\textcolor{black}{3.0} & \cellcolor[HTML]{C1DFCF}\textcolor{black}{25.5} & \cellcolor[HTML]{DEEEE5}\textcolor{black}{13.8} & \cellcolor[HTML]{F7FBF9}\textcolor{black}{3.4} & \cellcolor[HTML]{FEFFFE}\textcolor{black}{0.7} & \cellcolor[HTML]{FFFFFF}\textcolor{black}{0.3} & \cellcolor[HTML]{FFFFFF}\textcolor{black}{0.3} & \cellcolor[HTML]{FFFFFF}\textcolor{black}{0.3} & \cellcolor[HTML]{FFFFFF}\textcolor{black}{0.1} & \cellcolor[HTML]{FFFFFF}\textcolor{black}{0.0} \\
& 0.2--0.3 & \cellcolor[HTML]{FDFEFE}\textcolor{black}{0.9} & \cellcolor[HTML]{D3E8DD}\textcolor{black}{18.3} & \cellcolor[HTML]{CFE6D9}\textcolor{black}{20.0} & \cellcolor[HTML]{E6F2EC}\textcolor{black}{10.2} & \cellcolor[HTML]{F6FBF8}\textcolor{black}{3.5} & \cellcolor[HTML]{FDFEFE}\textcolor{black}{1.0} & \cellcolor[HTML]{FEFFFE}\textcolor{black}{0.5} & \cellcolor[HTML]{FFFFFF}\textcolor{black}{0.2} & \cellcolor[HTML]{FFFFFF}\textcolor{black}{0.1} & \cellcolor[HTML]{FFFFFF}\textcolor{black}{0.1} \\
& 0.3--0.4 & \cellcolor[HTML]{FDFEFE}\textcolor{black}{0.8} & \cellcolor[HTML]{E5F2EB}\textcolor{black}{10.7} & \cellcolor[HTML]{D4E9DE}\textcolor{black}{17.8} & \cellcolor[HTML]{D3E8DD}\textcolor{black}{18.0} & \cellcolor[HTML]{F1F8F4}\textcolor{black}{6.2} & \cellcolor[HTML]{FBFDFC}\textcolor{black}{1.8} & \cellcolor[HTML]{FEFFFE}\textcolor{black}{0.6} & \cellcolor[HTML]{FFFFFF}\textcolor{black}{0.2} & \cellcolor[HTML]{FFFFFF}\textcolor{black}{0.1} & \cellcolor[HTML]{FFFFFF}\textcolor{black}{0.0} \\
& 0.4--0.5 & \cellcolor[HTML]{FEFFFE}\textcolor{black}{0.4} & \cellcolor[HTML]{F5FAF7}\textcolor{black}{4.6} & \cellcolor[HTML]{E1EFE7}\textcolor{black}{12.5} & \cellcolor[HTML]{DAECE2}\textcolor{black}{15.5} & \cellcolor[HTML]{DFEEE6}\textcolor{black}{13.3} & \cellcolor[HTML]{F5FAF7}\textcolor{black}{4.7} & \cellcolor[HTML]{FCFEFD}\textcolor{black}{1.2} & \cellcolor[HTML]{FFFFFF}\textcolor{black}{0.4} & \cellcolor[HTML]{FFFFFF}\textcolor{black}{0.1} & \cellcolor[HTML]{FFFFFF}\textcolor{black}{0.0} \\
& 0.5--0.6 & \cellcolor[HTML]{FEFFFE}\textcolor{black}{0.5} & \cellcolor[HTML]{FDFEFE}\textcolor{black}{1.2} & \cellcolor[HTML]{F2F8F5}\textcolor{black}{5.5} & \cellcolor[HTML]{DAECE2}\textcolor{black}{15.5} & \cellcolor[HTML]{D4E9DE}\textcolor{black}{17.7} & \cellcolor[HTML]{E9F4EE}\textcolor{black}{9.2} & \cellcolor[HTML]{FBFDFC}\textcolor{black}{1.8} & \cellcolor[HTML]{FEFFFE}\textcolor{black}{0.5} & \cellcolor[HTML]{FFFFFF}\textcolor{black}{0.2} & \cellcolor[HTML]{FFFFFF}\textcolor{black}{0.1} \\
& 0.6--0.7 & \cellcolor[HTML]{FFFFFF}\textcolor{black}{0.3} & \cellcolor[HTML]{FCFEFD}\textcolor{black}{1.4} & \cellcolor[HTML]{F3F9F5}\textcolor{black}{5.2} & \cellcolor[HTML]{E7F3ED}\textcolor{black}{10.0} & \cellcolor[HTML]{D8EBE1}\textcolor{black}{16.3} & \cellcolor[HTML]{D8EBE1}\textcolor{black}{16.4} & \cellcolor[HTML]{EBF5F0}\textcolor{black}{8.6} & \cellcolor[HTML]{F9FCFB}\textcolor{black}{2.5} & \cellcolor[HTML]{FFFFFF}\textcolor{black}{0.3} & \cellcolor[HTML]{FFFFFF}\textcolor{black}{0.1} \\
& 0.7--0.8 & \cellcolor[HTML]{FFFFFF}\textcolor{black}{0.1} & \cellcolor[HTML]{FDFEFE}\textcolor{black}{0.9} & \cellcolor[HTML]{F7FBF9}\textcolor{black}{3.5} & \cellcolor[HTML]{E9F4EE}\textcolor{black}{9.3} & \cellcolor[HTML]{E0EFE7}\textcolor{black}{13.1} & \cellcolor[HTML]{D2E8DC}\textcolor{black}{18.7} & \cellcolor[HTML]{D7EAE0}\textcolor{black}{16.6} & \cellcolor[HTML]{EEF6F2}\textcolor{black}{7.1} & \cellcolor[HTML]{FBFDFC}\textcolor{black}{1.7} & \cellcolor[HTML]{FFFFFF}\textcolor{black}{0.1} \\
& 0.8--0.9 & \cellcolor[HTML]{FFFFFF}\textcolor{black}{0.2} & \cellcolor[HTML]{FDFEFE}\textcolor{black}{0.9} & \cellcolor[HTML]{FBFDFC}\textcolor{black}{1.8} & \cellcolor[HTML]{F7FBF9}\textcolor{black}{3.4} & \cellcolor[HTML]{E9F4EE}\textcolor{black}{9.1} & \cellcolor[HTML]{CFE6D9}\textcolor{black}{20.1} & \cellcolor[HTML]{C7E2D4}\textcolor{black}{23.2} & \cellcolor[HTML]{D6EADF}\textcolor{black}{16.8} & \cellcolor[HTML]{F0F7F3}\textcolor{black}{6.5} & \cellcolor[HTML]{FEFFFE}\textcolor{black}{0.7} \\
& 0.9--1.0 & \cellcolor[HTML]{FBFDFC}\textcolor{black}{1.9} & \cellcolor[HTML]{F6FAF8}\textcolor{black}{4.2} & \cellcolor[HTML]{F2F8F5}\textcolor{black}{5.5} & \cellcolor[HTML]{EFF7F2}\textcolor{black}{7.0} & \cellcolor[HTML]{D9EBE2}\textcolor{black}{15.8} & \cellcolor[HTML]{C1DFCF}\textcolor{black}{25.6} & \cellcolor[HTML]{90C5A9}\textcolor{black}{46.0} & \cellcolor[HTML]{52A479}\textcolor{white}{71.2} & \cellcolor[HTML]{248C55}\textcolor{white}{90.6} & \cellcolor[HTML]{108145}\textcolor{white}{98.7} \\
\bottomrule
& MAE & 0.04 & 0.15 & 0.19 & 0.21 & 0.24 & 0.25 & 0.23 & 0.19 & 0.13 & 0.03 \\
\bottomrule
\end{tabular}
}
    \label{tab:results:tc_validation}
\end{table}

\newpage
\section{Disturbance regimes in France}
\label{sec:appendix:disturbance_regimes}
\begin{figure}[H]
    \centering
    \includegraphics[width=0.8\linewidth]{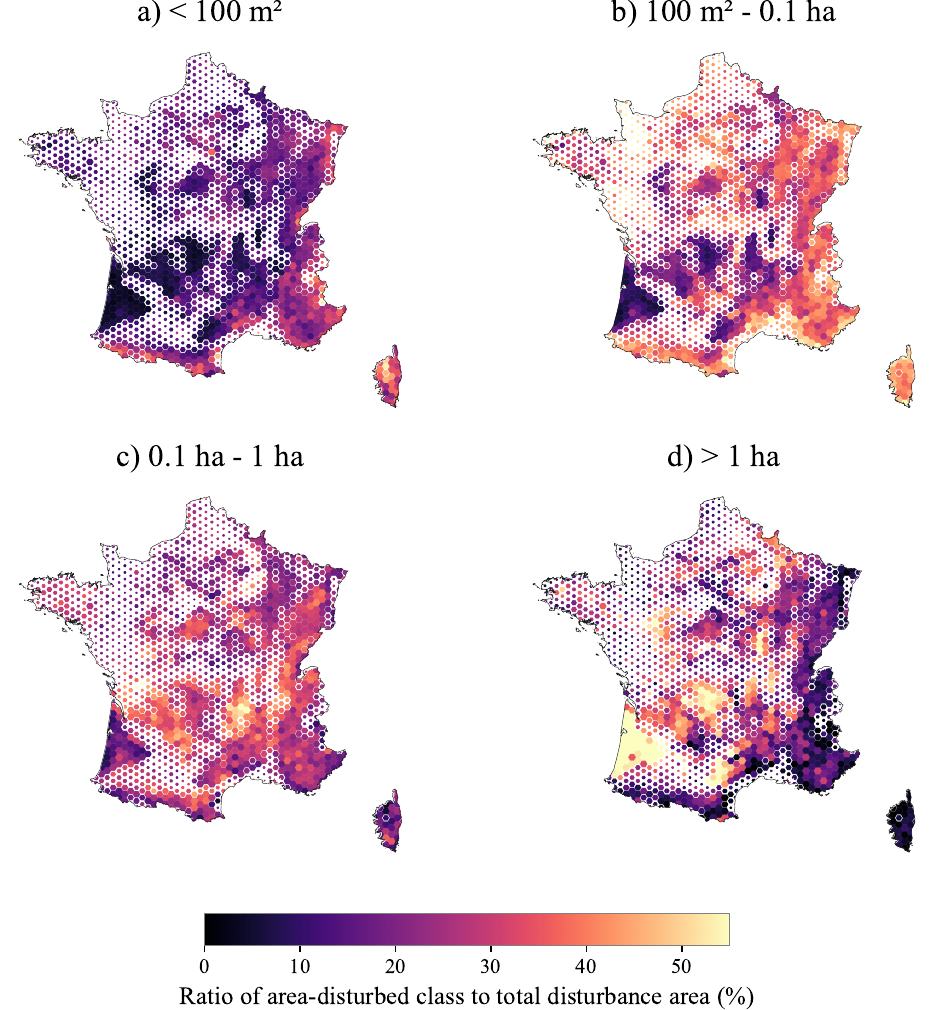}
    \caption{\textbf{Geographic distribution of \fdelta disturbance size class contributions across French forests (2015-2023).} Share of each size class in total disturbed area per hexagon, averaged over the 2015-2023 period: (a) <100\mtwo, (b) 100\mtwo-0.1~ha, (c) 0.1-1~ha, and (d) >1~ha. Brighter colors indicate a higher contribution of that size class to the total disturbed area. Hexagon size scales with forest fraction; hexagons are shown at full size when forest fraction exceeds 40\%. Together, the four panels characterize the dominant disturbance regime of each region, from fine-scale gap dynamics in mountain forests (a) to clearcut-dominated plantations in lowland areas (d).}
    \label{fig:disturbances_regimes_4_panels}
\end{figure}

\newpage
\begin{figure}[H]
    \centering
    \includegraphics[width=\linewidth]{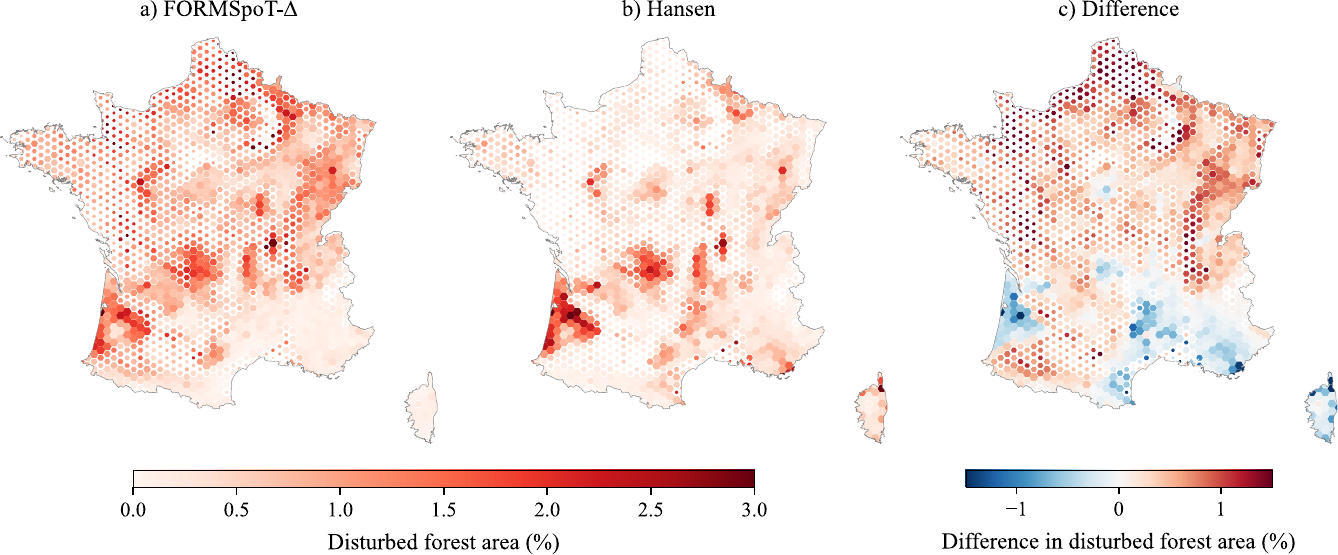}
    \caption{\textbf{Mean annual disturbed forest area fraction across France (2015-2023) and comparison with Hansen.} Fraction of forest area affected by disturbances per year, averaged over 2015-2023, derived from (a) \fdelta and (b) Hansen. (c) Difference between \fdelta and Hansen disturbed area fractions; red hexagons indicate locations where \fdelta detects more disturbed area than Hansen, blue hexagons where Hansen detects more. Hexagon size scales with forest fraction; hexagons are shown at full size when forest fraction exceeds 40\%.}
    \label{fig:appendix_hanscen_comparison}
\end{figure}

\section{Published paper and Supplementary materials}
The paper published in \textit{Remote Sensing of Environment} and the supplementary materials can be accessed at \url{https://doi.org/10.1016/j.rse.2026.115631}.

\small

\bibliographystyle{elsarticle-harv}
\bibliography{Review_round_2/references_review_2}

\end{document}